\documentclass[10pt,journal,compsoc]{IEEEtran}

\usepackage{amsfonts}
\usepackage[linesnumbered,ruled,vlined]{algorithm2e}
\usepackage{threeparttable}
\usepackage{amsmath}

\usepackage{cuted}

\usepackage{tikz}
\newcommand*\emptycirc[1][1ex]{\tikz\draw[thick] (0,0) circle (#1);} 

\newcommand*\fullcirc[1][1ex]{\tikz\fill (0,0) circle (#1);}

\usepackage{hyperref}
\hypersetup{
	colorlinks=true,
	linkcolor=black,
	filecolor=magenta,      
	urlcolor=black,
	citecolor=black
}

\usepackage[nocompress]{cite}
\usepackage{booktabs}
\usepackage[psamsfonts]{amssymb}
\usepackage{ragged2e}
\usepackage{balance}

\usepackage{multirow}

\usepackage{xcolor}

\newcommand{\hlt}{\textcolor{black}}

\newcommand{\hltt}{\textcolor{black}}

\ifCLASSOPTIONcompsoc
  \usepackage[nocompress]{cite}
\else
  \usepackage{cite}
\fi

\ifCLASSINFOpdf
\else
\fi

\begin{document}

\title{Projective Ranking-based GNN Evasion Attacks}

\author{He Zhang, Xingliang Yuan, Chuan Zhou, Shirui Pan
\IEEEcompsocitemizethanks{\IEEEcompsocthanksitem H. Zhang and 
X. Yuan are with the Department of Software Systems and Cybersecurity, Faculty of IT, Monash University, Clayton, VIC 3800, Australia. 
E-mail: \{he.zhang1, xingliang.yuan\}@monash.edu ;
\IEEEcompsocthanksitem S. Pan is with School of Information and Communication Technology, Griffith University, Australia. 
Email: s.pan@griffith.edu.au;
\IEEEcompsocthanksitem C. Zhou is with Academy of Mathematics and Systems Science, Chinese Academy of Sciences, China. 
Email: zhouchuan@amss.ac.cn;
\IEEEcompsocthanksitem Corresponding Author: Xingliang Yuan and Shirui Pan.}
}
%
%

\markboth{Accepted by IEEE Transactions on Knowledge and Data Engineering, DOI:10.1109/TKDE.2022.3219209}%
{Shell \MakeLowercase{\textit{et al.}}: Bare Demo of IEEEtran.cls for Computer Society Journals}
%



\IEEEtitleabstractindextext{%
\begin{abstract}
\justifying
 Graph neural networks (GNNs) offer promising learning methods for graph-related tasks. However, GNNs are at risk of adversarial attacks. Two primary limitations of the current evasion attack methods are highlighted: (1) The current \textit{GradArgmax} ignores the ``long-term" benefit of the perturbation. It is faced with zero-gradient and invalid benefit estimates in certain situations. (2) In reinforcement learning-based attack methods, the learned attack strategies might not be transferable when the attack budget changes. To this end, we first formulate the perturbation space and propose an evaluation framework and the projective ranking method. We aim to learn a powerful attack strategy then adapt it as little as possible to generate adversarial samples under dynamic budget settings. In our method, based on mutual information, we rank and assess the attack benefits of each perturbation for an effective attack strategy. By projecting the strategy, our method dramatically minimizes the cost of learning a new attack strategy when the attack budget changes. In the comparative assessment with \textit{GradArgmax} and \textit{RL-S2V}, the results show our method owns high attack performance and effective transferability. The visualization of our method also reveals various attack patterns in the generation of adversarial samples.

\end{abstract}

\begin{IEEEkeywords}
Adversarial attacks, graph neural networks, graph classification.
\end{IEEEkeywords}}

\maketitle

\IEEEdisplaynontitleabstractindextext

%
\IEEEpeerreviewmaketitle

\IEEEraisesectionheading{\section{Introduction}
\label{sec:introduction}}

\IEEEPARstart{G}{raphs} consist of nodes and edges defined between these nodes. As an abstract data type, graphs have powerful modelling capabilities. By extracting attributes from entities and describing their relationships, graphs can represent a range of objects or technical systems in real-world situations,  
such as drug molecules \cite{ma2018drug}, biological networks \cite{fortunato2018science}, and traffic networks \cite{zheng2021spatio,jin2022multivariate}. Graph neural networks (GNNs) have proven to be effective graph learning methods \cite{zheng2022graph} for exploring graph data and have demonstrated promising performance on node classification \cite{jin2021multi,wu2020openwgl}, link prediction \cite{xiong2021semi}, anomaly detection \cite{liu2021anomaly,zheng2022unsupervised}, and graph classification \cite{Gao2021tkde}. For example, in biological chemistry, the GNNs are engaged in recognizing the chemical properties of molecules.


In practice, GNNs raise urgent security concerns, although they have garnered considerable attention in the context of complex graph-structured data applications \cite{Liu2021tkde,DBLP:conf/nips/LiWWL20,liu2019shifu2,li021tkde}. Generally, graph neural networks are regarded as the generalization of deep neural networks into graph data. However, GNNs inherit the vulnerability of deep neural networks despite possessing adequate expressive power. Several recent studies demonstrate that adversaries can attack GNNs \cite{jin2020adversarial, chen2020survey,WuYPY21, sun2018adversarial,  WuYPY22, zhang2022robust}. Among others, evasion attacks~\cite{DBLP:conf/icml/DaiLTHWZS18,DBLP:conf/kdd/ZugnerAG18} are notoriously dangerous.
Attackers could perturb test samples to generate adversarial samples of a victim model trained on original clean data. On adversarial samples, the victim model will give incorrect results, different from those of clean samples.

Prior work on GNN evasion attacks has demonstrated that well-designed slight perturbations are able to significantly degrade the performance of the victim models \cite{DBLP:conf/icml/DaiLTHWZS18}. This type of attack is tricky since adversarial samples are similar to clean samples in appearance \cite{chen2020survey}. Adding a few elaborate adversarial edges can dramatically reduce the classification accuracy of the victim models \cite{DBLP:conf/kdd/ZugnerAG18}. To create an adversarial example, attackers can change the features of few nodes or add/remove a few edges in a clean graph. A \textit{budget} $k$ with positive integer values controls the number of these perturbation operations to make their perturbations stealthy. Moreover, once the models are deployed, it is practical for attackers to launch the evasion attacks at any time \cite{sun2018adversarial}.

The existing studies of evasion attacks on the GNN models mainly focus on node classification and link prediction, while few of them are designed for graph classification \cite{chen2020survey,sun2018adversarial,jin2020adversarial}. We note there are some \textit{limitations} in those methods for graph classification, which need to be solved to improve the attack practicality. 
\begin{itemize}
    \item In the vanilla gradient-based attack method \textit{GradArgmax} \cite{DBLP:conf/icml/DaiLTHWZS18}, the greedy mechanism uses the gradient on the graph in the perturbation process to make the perturbation decision. Although the gradient information could approximate the attack benefit to some extent, it ignores the long-term benefit of operation at each step \cite{DBLP:conf/icml/DaiLTHWZS18}. Moreover, \textit{GradArgmax} is not efficient since it needs the up-to-date gradient information to complete each step of all the perturbation operations.
    \item Attack methods \cite{DBLP:conf/icml/DaiLTHWZS18,ma2019attacking} use reinforcement learning (RL) to perturb clean samples and use the final predicted result on adversarial samples to estimate the attack benefit. When the perturbation budget changes, they have to be retrained to learn an attack strategy suitable for the new budget. In addition, we observe that \textit{RL-S2V} \cite{DBLP:conf/icml/DaiLTHWZS18}, a known attack method, does not appear to be able to generate adversarial samples in high attack success rates when the perturbation budget increases~\cite{ma2019attacking}.
\end{itemize}

Motivated by these observations, we summarize the challenges in building an effective evasion attack against GNNs for graph classification: (1) \textit{in the generation of powerful adversarial samples, how to measure the attack benefits of each operation based on the clean graph}, and (2) \textit{how to adapt the learned attack strategy to changed perturbation settings (e.g., budget) with least effort}.

To address these challenges, we first formulate the perturbation space composed of all possible perturbation operations and present four properties of it. Then, we propose an evaluation framework consisting of three principles decomposed from the adversary's expectation. We use these three principles as the design goal of our method and employ them to evaluate typical evasion attack methods. In our evaluation framework, we consider a basic fact in the evasion attack: an intelligent attacker always expects to choose the perturbation operation which will bring the most significant attack benefit at each perturbation step. This fact includes three critical principles for evasion attack: {\itshape attack benefit}, {\itshape operation ranking} and {\itshape baseline graph}. Finally, we propose the projective ranking method as an instantiation under the evaluation framework. 

In our method, we expect to find a suitable metric that considers the attack benefits of each operation to rank them and then use this ranking directly to generate adversarial samples under different perturbation budget settings. Toward the first challenge, we employ the mutual information (MI) between elements of perturbation space and attackers' goals as a measure to evaluate the importance of each possible operation. We regard the perturbation space ranking as the learned attack strategy. Inspired by the projected gradient descent method, we propose to project the learned attack strategy into practical attackers' perturbations with the specified budget for the second challenge. Through this projection operation, the adjustment cost of the learned attack strategy is almost zero.

Our contributions are summarized below:
\begin{itemize}
    \item We formulate the perturbation space and its four properties, and we propose an evaluation framework for evasion attack methods and analyze typical methods from the principles of the framework.
    \item We propose the projective ranking method. We first employ the mutual information to measure the attack benefits of a perturbation operation towards strong attack performance, then use the projection operation to reduce the cost of the strategy adaptation process by considering the transferability of attack strategies.
    \item We conduct evaluations on several real world datasets and a synthesized dataset, and our attack method achieves high attack performance and effective transferability. The visualization of adversarial samples present various attack patterns and reveals the vulnerability of the victim model.
\end{itemize}

\begin{table*}[t]
\centering
\caption{Evasion Attack Methods for Graph Classification.}
\vspace{-5pt}
\label{tab:relatedwork}
\begin{threeparttable}
\renewcommand{\arraystretch}{1.1}
\begin{tabular}{|c|c|c|c|c|c|c|}
\hline
 &  & \multicolumn{2}{c|}{Operations} &  & \multicolumn{2}{c|}{Transferability   of Attack Strategy} \\ \cline{3-4} \cline{6-7} 
\multirow{-2}{*}{Method} & \multirow{-2}{*}{Decision-making information} & Edges & Features & \multirow{-2}{*}{Attack Goal} & Unseen   samples & Budget \\ \hline
Rand\tnote{*} & Random & Add/Remove & - & Untargeted   attack & - & - \\ \hline
GradArgmax & Gradient   of loss function & Add/Remove & - & Untargeted   attack & - & - \\ \hline
RL-S2V & Reward in RL & Add & - & Untargeted/Targeted   attack & \checkmark & - \\ \hline
Rewatt & Reward in RL & Rewrite & - & Untargeted/Targeted   attack & \checkmark & - \\ \hline
HGP\tnote{*} & Gradient   of selection function & Add/Remove & modify & Untargeted   attack & - & - \\ \hline
Ours & Mutual   information & Add & - & Untargeted/Targeted   attack & \checkmark & \checkmark \\ \hline
\end{tabular}
\begin{tablenotes}
    \footnotesize \item[*] Rand means the \textit{RandomSampling}, HGP means the method for attacking the pooling operation in Hierarchical
Graph Pooling Neural Networks.
\end{tablenotes}
\end{threeparttable}
\vspace{-10pt}
\end{table*}

In this paper, we review different attack methods for graph classification in Section \ref{sec:relatedwork} and introduce the graph classification and evasion attacks in Section \ref{sec:preliminary}. In Section \ref{sec:method}, we first formulate the perturbation space and show its four properties. Then we propose the evaluation framework of evasion attack methods and show the instantiation design of our method. Next, we show the details of the projective ranking method. In Section \ref{sec:frameworks_evaluation}, we use the evaluation framework to make a comparison evaluation of typical methods. In Section \ref{sec:experiments}, we experimentally demonstrate the high attack performance and effective transferability of our method, and the attack patterns in our adversarial samples reveal the weaknesses of the victim model. Finally, we show the conclusion and discuss the future works in Section \ref{conclusion}.

\section{Related Work}
\label{sec:relatedwork}
Current research shows that graph neural networks are vulnerable to adversarial attacks. According to the intention of attackers, the adversarial attack methods for graph classification include evasion attacks, backdoor attacks, and poisoning attacks. 
\subsection{Evasion Attack Methods on GNNs}
Evasion attacks perturb the inputs of the GNNs during their inference phase and intend to degrade the performance of the victim model. Hence, the more the classification accuracy of the victim model decreases, the better the performance of an attack method. As the most intuitive approach, \textit{RandomSampling} randomly perturbs the graph structure or nodes' features of clean graphs to generate adversarial samples. However, the attack performance of this method is lower than adding well-designed perturbations. In the {\itshape GradArgmax} method, the attackers own the ability to access the victim model. They first calculate the gradient information of the loss function on each possible perturbation operation, then use the gradient information to choose the perturbation operation that maximizes the loss of the victim model. 

To better describe the perturbations on graph data, Dai et al. \cite{DBLP:conf/icml/DaiLTHWZS18} proposed to employ the Markov decision process to model the whole perturbation process and use reinforcement learning to obtain an attack strategy on current clean samples. The graph data in the perturbation process is modeled as the state of reinforcement learning. The action space comprises all perturbation operations in the current state, and the final attack result defines the reward function. Benefiting from the modelling and learning ability of reinforcement learning, the attack performance of {\itshape RL-S2V} outperforms that of {\itshape RandSampling} and {\itshape GradArgmax} on synthetic data \cite{DBLP:conf/icml/DaiLTHWZS18}. 

To improve the stealthiness of attack, Ma et al. proposed {\itshape ReWatt} to redefine the action space of reinforcement learning \cite{ma2019attacking}. The perturbations generated by {\itshape ReWatt} are more unnoticeable, as the edges number of adversarial samples is equal to those of clean samples. Besides the above attack methods, attackers could also attack the components of GNNs if they know a specified operation is included in the victim model. For example, some methods introduce hierarchical pooling operations in the models, which select typical nodes to compress the node number and determine the structure of the coarse graph \cite{ying2018hierarchical,ma2019graph,zhang2019hierarchical}. 

Tang et al. proposed to attack the selecting operation in pooling operation \cite{tang2020adversarial}. By training a surrogate model with hierarchical pooling on some clean data, the attackers obtain the selecting function that simulates the victim model's hierarchical pooling. Based on this selection function, the attackers could obtain the adversarial samples that invalidate the pooling in the victim model. However, this method may not be available for attacking models that employ global pooling operations. A comparison of representative evasion attack methods is summarized in Table \ref{tab:relatedwork}.
\subsection{Differences in Evasion Attacks}
In addition to the differences in technical methods, the evasion attack methods also have differences in perturbation space, the definition of imperceptible perturbations, the attacker's knowledge, and the attack goal. In the aspect of perturbation space, the attackers usually make perturbations by adding fake nodes with the fake features \cite{wang2018attack}, modifying node features \cite{DBLP:conf/ijcai/Wu0TDLZ19}, adding or deleting edges \cite{DBLP:conf/icml/BojchevskiG19}. In graph classification evasion attacks, {\itshape ReWatt}, {\itshape RL-S2V}, {\itshape RandSampling} and {\itshape GradArgmax} only modify graph structure. The attack method in \cite{tang2020adversarial} makes perturbations in node features or graph structure. 

In the aspect of imperceptible perturbations, the distance between the original clean graph and the modified adversarial graph is usually limited by the perturbation budget, which is defined as the number of nodes modified or distance of feature vectors, or the number of edges modified by adding/deleting/rewriting. In the aspect of attackers' knowledge, based on how much information an attacker knows about the victim model, the attack methods are generally divided into white-box, grey-box, and black-box attack. A white-box attack means the attackers can access all information about the victim model like architecture, parameters, training input, and labels. A grey-box attack indicates that only limited information about the victim model, like training data labels, is available. As the strictest setting, the black-box attack only allows attackers to do queries of samples for output or labels \cite{jin2020adversarial}. 

In graph classification evasion attack, {\itshape GradArgmax} is a white-box attack method, {\itshape RL-S2V} and {\itshape ReWatt} are typical black-box attack methods based on reinforcement learning. In the aspect of the attacker's goal, the attacks are divided into untargeted attacks and targeted attacks. In an untargeted attack, the attackers expect the victim model to classify the adversarial samples into labels different from their original labels. As a more strict attack goal, targeted attacks attempt to fool the victim model by specifying the output categories of the adversarial samples. 

\subsection{Poisoning and Backdoor Attacks on GNNs}
Besides evasion attacks, another two typical adversarial attacks on GNN models are poisoning attacks and backdoor attacks. The poisoning attacks \cite{fang2018poisoning} suppose the attackers own the ability to poison training data and label. In this way, the performance of GNN models trained on poisoned data will degrade dramatically on clean samples. In backdoor attacks \cite{zhang2020backdoor,xi2021graph}, the attackers inject a fixed or adaptive trigger into clean training data and change their labels to the desired categories. As a result, the models trained on these data performs well on the clean samples but predicts the desired labels once the well-designed trigger is injected into the clean samples.

\section{Preliminary}
\label{sec:preliminary}
\subsection{Graph Classification}
Assuming \begin{math}\mathnormal{G}=\{\mathcal{V}, \mathcal{E}\}\end{math} is a graph, where $\mathcal{V}=\left\{v_{1}, \ldots, v_{|\mathcal{V}|}\right\}$ is the set of nodes, \begin{math}\mathcal{E}=\left\{e_{1}, \ldots, e_{|\mathcal{E}|}\right\}\end{math} is the set of edges of graph $G$. The edges set $\mathcal{E}$ describes the structural information of $G$. It can also be expressed as the adjacency matrix $\mathnormal{A}\in\{0,1\}^{|\mathcal{V}| \times|\mathcal{V}|}$, where $\mathnormal{A}_{ij}=1$ means the existence of the edge from $v_{i}$ to $v_{j}$, otherwise $\mathnormal{A_{ij}}=0$. The features associated with nodes are expressed as matrix $\mathnormal{X}\in\mathbb{R}^{|\mathcal{V}| \times d}$, where the $i$-th row of $\mathnormal{X}$ is the features of node $\mathnormal{v_{i}}$ and $d$ is the dimension of features. So a graph can also be expressed as $\mathnormal{G}=\{\mathnormal{A}, \mathnormal{X}\}$. 

In the graph classification, a set of graphs is denoted by $\mathcal{G}=\left\{G_{i}\right\}_{i=1}^{N}$ and a label $y_{i}\in\mathcal{Y}=\{1,2,\ldots,Y\}$ is associated with each graph $G_{i}$, where $N$ is the number of graphs and $Y$ is number of categories. The dataset $\mathcal{D}=\left\{\left(G_{i},y_{i}\right) \right\}_{i=1}^{N}$ is composed of pairs of graph and its label. In graph learning, the classifier $\mathnormal{f}\in\mathcal{F}:\mathcal{G}\rightarrow\mathcal{Y}$ is trained and expected to learn the mapping from graph $\mathnormal{G}$ to its label $\mathnormal{y}$ with optimal parameters $\theta$ that minimize the below loss function:
\begin{equation}
    \mathcal{L}=\frac{1}{N} \sum_{i=1}^{N} L\left(f_\theta \left(G_{i}\right), y_{i}\right),
\end{equation}
where $\mathnormal{L}(\cdot, \cdot)$ is used to measure the distance between the predicted and ground-truth labels. A general instance of $L(\cdot, \cdot)$ is the cross-entropy function.

\subsection{Graph Neural Networks Model}
The graph neural networks (GNNs) is a family of architectures of neural networks that are designed to process graph data $G=\{\mathcal{V}, \mathcal{E}\}$. These models iteratively update the expression of nodes by message passing and aggregation as below \cite{gilmer2017neural}:
\begin{align}
\label{gnnmessage}
m_{v}^{t+1}&=\sum_{w\in\mathcal{N}(v)}M_{t}\left(h_v^t,h_w^t,e_{vw}\right),\\\label{gnnupdating}
h_v^{t+1} &= U_t\left(h_v^t,m_v^{t+1}\right),
\end{align}
where $N(v)$ is the set of neighbours of node $v$ in graph $G$, $h_v^{t}$ means the hidden expression of node $v$ at time $t \in \{1,2,\ldots,T\}$ and $e_{vw}$ is the features of the edge from node $v$ to node $w$. $M_t(\cdot,\cdot,\cdot)$ is the message function and $U_t(\cdot,\cdot)$ is the vertex expression updating function. After the message passing phase, the expression of whole graph $G$ is obtained by the readout operation:
\begin{equation}
    h_{G}=R\left(\left\{h_{v}^{T} \mid v \in G\right\}\right)
\end{equation}
where $R(\cdot)$ is the readout operation, and it is invariant to
permutations of the nodes. A general instance of $R(\cdot)$ is the \textit{max-pooling} or \textit{sum-pooling}. 

In graph classification, the model $f_{\theta}$ based on the above architecture is generally trained under an inductive learning setting, in which the classifier learns on training data $\mathcal{D}_{train}$ and then make graph label prediction on test data $\mathcal{D}_{test}$.

\subsection{Problem Formulation}
\label{sec:prob_formulate}
Assuming a GNN model $f_{\theta}$ is trained on clean graph samples and then employed to predict the category of graph data $\mathcal{D} = \{G_j\}_{1}^M$. 
In evasion attacks, an attacker $\mathcal{T}$ attempts to make unnoticeable perturbations on the original $G$ to degrade the performance of this victim model $f_{\theta}$. 
We use $\mathcal{T}_{f}(G)$ to indicate the attacker's perturbation on $G$ which is specifically designed for the classifier $f_{\theta}$. In untargeted evasion attacks, the attackers expect the predicted category of an adversarial sample is different from its true category. More precisely, the objective of {\itshape untargeted evasion attacks} on the victim model $f_{\theta}$ is 
\begin{equation}
\label{equ:optimization_uta}
\begin{array}{ll}
\max_{\widehat{G}} &  \sum_{j=1}^{M} \mathbb{I}\left(f_{\theta}(\widehat{G}_{j}) \neq y_j\right) \\
\mathnormal { s.t. } & \widehat{G}=\mathcal{T}_{f} (G)\\
& \mathcal{I}(\widehat{G}, G;k)=1.
\end{array}
\end{equation}
Here 
$\widehat{G}$ is the adversarial sample generated from the clean sample $G$, and $\mathbb{I}(\cdot)$ is the binary indicator function. \hlt{$k$ is the perturbation budget of $\mathcal{T}_{f}$. Given a specific $k$, $\mathcal{I}(\cdot, \cdot;k)$ is a similarity measure function whose output is 1 when two input graph samples are semantically the same and 0 otherwise.} Particularly, in targeted evasion attacks, attackers expect the predicted category to be a specified category $y_t$, which is different from its true category $y_j$. The objective of {\itshape targeted evasion attacks} is 
\begin{equation}
\label{equ:optimization_ta}
\begin{array}{ll}
\max_{\widehat{G}} &  \sum_{j=1}^{M} \mathbb{I}\left(f_{\theta}(\widehat{G}_{j}) = y_t\right) \\
\mathnormal{ s.t. } & \widehat{G}=\mathcal{T}_{f} (G)\\
& \mathcal{I}(\widehat{G}, G; k)=1,\\
& y_t\neq y_j.
\end{array}
\end{equation}

\noindent
\hlt{\textit{Note.} 
The probability function $p(\cdot)$ can be employed as a substitute of $\mathbb{I}(\cdot)$ to describe objective functions \cite{jin2020adversarial}.
Given $G_{j}$ and budget $k$, we use
$O(\widehat{G}_j;k)$ 
to denote the value of objective function in equation (\ref{equ:optimization_uta})/(\ref{equ:optimization_ta}) when using up all budget $k$ (e.g., in equation (\ref{equ:optimization_uta}), $O(\widehat{G}_j;k)=\mathbb{I}\left(f_{\theta}(\widehat{G}_{j}) \neq y_j\right)$ or $p\left(f_{\theta}(\widehat{G}_{j}) \neq y_j\right)$).} 

\section{Proposed Approach}
\label{sec:method}
Our projective ranking method mainly includes the ranking module and the projection module. In this section, we first formulate the perturbation space and propose the principles of our evaluation framework. Then we present the detail of the ranking module and the perturbation projection operation. Finally, we give an algorithm to show the generation of adversarial samples. 

\subsection{Attack Setting}

In our method, given a victim model $\mathnormal{f_{\theta}}$, we assume that the attackers can have access to the node embedding and predictive probability distribution of $\mathnormal{f_{\theta}}$
to learn the attack strategy. Then, only accessing the embedding of nodes, the attackers may utilize the learned strategy to attack any samples of the victim model under any budget setting.   

To further explore the transferability of the learned strategy (in Section \ref{sec:transferability_data_model}), we assume an intelligent attacker attempts to fool the victim models by accessing the nodes embedding of the target models based on the attack strategy learned from the other source models engaged in the similar data domain. We note that the attacker does not know any other knowledge (like the architecture or parameters of the target models) except for the embedding of nodes. This scenario reflects the vulnerability of the models that use the pre-training models \cite{DBLP:conf/iclr/HuLGZLPL20}. 


\subsection{Perturbation Space}
\label{sec:perturbationspace}
In an evasion attack, an attacker $\mathcal{T}$ can perturb a clean sample $G$ to obtain an adversarial sample $\widehat{G}$. We formulate the process of this perturbation on graph $G$ as
\begin{equation}
    \widehat{G} = \mathcal{T}_{f}(G) = G + \Delta G,
\end{equation}
where $\Delta G = \{\Delta A, \Delta X\}$ is the perturbation graph. To be more precise, the graph structure and node features of $\widehat{G}$ are expressed as
\begin{equation}
\begin{aligned}
    \{\widehat{{A}}, \widehat{{X}}\} &= \{{A} + \Delta{A}, {X} + \Delta{X}\}, \\
     \Delta{A} &= {m}_{{A}}\odot \left[ \mathbb{I}(add)\cdot \left( {I}_{{A}}-{A}\right) +  \mathbb{I}(del)\cdot \left(-{A}\right) \right],\\
     \Delta{X} &= {m}_{{X}}\odot ({I}_{{X}}-2{X}),
\end{aligned}
\end{equation}
where $\mathnormal{A}\in\{0,1\}^{|\mathcal{V}| \times|\mathcal{V}|}$ is the adjacency matrix, $X\in\{0,1\}^{|\mathcal{V}|\times d}$ is binary node features matrix, ${I}_{{A}} = J_{|\mathcal{V}|}-I_{|\mathcal{V}|}$, ${I}_{{X}} = J_{|\mathcal{V}|,d}$ ($J$ is the all-one matrix, $I$ is the identity matrix). $|\mathcal{V}|$ is the number of nodes in $G$, and $d$ is the size of node features. $m_{A} \in \{0,1\}^{|\mathcal{V}| \times|\mathcal{V}|}$ and $m_{X} \in \{0,1\}^{|\mathcal{V}| \times d}$ represent the masks of graph structure and node features, respectively. $\mathbb{I}(add)$/$\mathbb{I}(del)$ is the indicator function to show if adding/deleting edge operation is allowed in generating adversarial samples. $\odot$ is the Hadamard product and $\cdot$ is the scalar multiplication. Therefore, the perturbation graph $\Delta G = \{\Delta A, \Delta X\}$ is actually defined by the combination of specified structure mask $m_{A}$ and feature mask $m_{X}$. The similarity measure function in (\ref{equ:optimization_uta}) and (\ref{equ:optimization_ta}) is refined as 
\begin{equation}
    \mathcal{I}(\widehat{G},G) = \mathbb{I}\left(a|| {m}_{{A}} ||_{1} + || {m}_{{X}}||_{1} \leq k \right),
\end{equation}
where $||\cdot||_{1}$ is 
the L1 Norm, $k$ is the budget of perturbations. $a$ is a scalar coefficient, and it is $1$ if $G$ is a directed graph and $\frac{1}{2}$ otherwise, since the adjacency matrix of the undirected graph is symmetric.

Given the clean sample $G$, the {\itshape perturbation space} $\mathbb{T}(G)$ of $G$ is defined as the set of all possible perturbation graph $\Delta G =\{\Delta A, \Delta X\}$. The {\itshape size} of a perturbation graph $\Delta G$ is defined as:
\begin{equation}
    ||\Delta G||= a|| {m}_{{A}} ||_{1} + || {m}_{{X}}||_{1} \in \mathbb{Z}_{0}^{+},
\end{equation}
where $\mathbb{Z}_{0}^{+}$ is the set of non-negative integer. In the {\itshape perturbation space} $\mathbb{T}(G)$, the operators $\odot$ and $+$ on perturbation graphs are defined as:
\begin{align*}
\Delta G_{\alpha} \odot \Delta G_{\beta} &= \{\Delta A_{\alpha}\odot\Delta A_{\beta}, \Delta X_{\alpha}\odot\Delta X_{\beta}\}, \\
\Delta G_{\alpha} + \Delta G_{\beta} &= \{{}_{\llcorner}^{\ulcorner}\Delta A_{\alpha}+\Delta A_{\beta} {}_{\lrcorner}^{\urcorner}, {}_{\llcorner}^{\ulcorner}\Delta X_{\alpha}+\Delta X_{\beta}{}_{\lrcorner}^{\urcorner}\},
\end{align*}
where ${}_{\llcorner}^{\ulcorner}\cdot{}_{\lrcorner}^{\urcorner}$ means the elements of the input matrix is limited in $\lbrack-1,1\rbrack$. 
The {\itshape perturbation space} $\mathbb{T}(G)$ with budget $k$ is defined as
\begin{equation}
    \mathbb{T}_{k}(G) = \{\Delta G \ \big|\ ||\Delta G||=k\}.
\end{equation}
Specially, we call $\Delta G \in \mathbb{T}_{1}(G)$ an {\itshape operation element} of {\itshape perturbation space} $\mathbb{T}(G)$. 

Given the above definitions and operators, the \textit{perturbation space} $\mathbb{T}(G)$ owns the following properties:

\vspace{1mm}
\noindent
\textbf{\textit{Property 1.}} If $\Delta G_{0} \in \mathbb{T}_{0}(G)$, then $\forall\ \Delta G_{\alpha} \in \mathbb{T}_{m}(G)$, we have
\begin{equation}
\label{Property1}
\begin{split}
\Delta G_{0}\odot\Delta G_{\alpha}=\Delta G_{0},\\
\Delta G_{0}+\Delta G_{\alpha}=\Delta G_{\alpha}.
\end{split}
\end{equation}


\noindent
\hlt{
\textbf{\textit{Property 2.}} Given $\Delta G_{\alpha}$, $\Delta G_{\beta} \in \mathbb{T}_{1}(G)$, $\Delta G_{\alpha}\odot\Delta G_{\beta} \in \mathbb{T}_{1}(G)$ if $\alpha = \beta$, otherwise $\Delta G_{\alpha}\odot\Delta G_{\beta} \in \mathbb{T}_{0}(G)$.
}
\\
\noindent
\hlt{\textbf{\textit{Property 3.}} Given $\Delta G_{\alpha} \in \mathbb{T}_{1}(G)$, $\Delta G_{\beta} \in \mathbb{T}_{\kappa-1}(G)$ ($\kappa \in \mathbb{Z}^{+}$), $\Delta G_{\alpha} + \Delta G_{\beta} \in \mathbb{T}_{\kappa}(G)$ if $\Delta G_{\alpha}\odot\Delta G_{\beta} \in \mathbb{T}_{0}(G)$.
}
\vspace{1mm}
\noindent
\hlt{\textbf{\textit{Property 4.}} $\forall\ \Delta G_{\alpha} \in \mathbb{T}_{m}(G)$, $\Delta G_{\beta} \in \mathbb{T}_{n}(G)$, we have $\Delta G_{\alpha} + \Delta G_{\beta} \in \mathbb{T}_{m+n-||\Delta G_{\alpha} \odot \Delta G_{\beta}||}(G)$.
}

\noindent
\hltt{
\textit{Proof}: Given the definition of $\mathbb{T}_{k}(G)$, the result is a direct consequence of the following calculations. Based on the definitions of $+$ and $\odot$ operation, we have 
\begin{equation*}
\begin{split}
    ||\Delta G_{\alpha} + \Delta G_{\beta}||&=||\Delta G_{\alpha}||+||\Delta G_{\beta}||-||\Delta G_{\alpha} \odot \Delta G_{\beta}||\\
    &= m+n - ||\Delta G_{\alpha} \odot \Delta G_{\beta}||
\end{split}
\end{equation*}}

Based on property 3, it is easy to obtain that the {\itshape perturbation space} $\mathbb{T}_{k}(G)$ is actually composed of $k$ different {\itshape operation elements} of $\mathbb{T}(G)$, which means that
\hlt{
\begin{equation}
\label{TkGComposition}
\begin{split}
    \mathbb{T}_{k}(G)=\{\Delta G\ \big|&\ \Delta G=\sum_{\kappa=1}^{k}\Delta G_{\kappa};
    \Delta G_{\kappa}\in\mathbb{T}_{1}(G),\\ & 
    \Delta G_{\kappa_{i}}\odot\Delta G_{\kappa_{j}} \in \mathbb{T}_{0}(G),
    \ \forall \kappa_{i}\neq\kappa_{j}\},
\end{split}
\end{equation}
}
and the {\itshape size} of $\mathbb{T}_{k}(G)$ is
\begin{equation}
    \big|\mathbb{T}_{k}(G)\big|= C_{E}^{k},\ E=\big|\mathbb{T}_{1}(G)\big|,
\end{equation}
where $|\cdot|$ is the size of a set, and $C$ is the combination operation of two integers.

\subsection{Projective Ranking}
\subsubsection{Design Goals and Principles in Evaluation Framework}
\label{sec:evaluationframe}




\hltt{Incorporating with the objective in (\ref{equ:optimization_uta})/(\ref{equ:optimization_ta}), we also consider a fundamental expectation of the adversary as a } \hlt{ \textbf{design goal} of evasion attack methods: \textit{An intelligent attacker always expects to choose the perturbation operation which will bring the most significant attack benefit at each perturbation step for achieving stealthy and effective attacks.}}

\hltt{Assuming that $O(\widehat{G};k)$ (see Section \ref{sec:prob_formulate}) results from the collaboration of $k$ different $\Delta G_{i}$,} 
\hlt{where $\Delta G_{i} \in \mathbb{T}_{1}(G)$ and $\sum_{i=1}^{k}\Delta G_{i}\in \mathbb{T}_{k}(G)$. $B(\cdot)$ is the attack benefit function defined on $\Delta G_{i}$, $B(\Delta G_{i})$ presents the contribution of $\Delta G_{i}$ to $O(\widehat{G};k)$. 
\hltt{The intelligent attackers expect to ranking these k different $\Delta G_{i}$ w.r.t $B(\Delta G_{i})$, and at each perturbation step they choose} $arg\max_{\Delta G_i} B(\Delta G_i)$.}
\hltt{Moreover, the objective in (\ref{equ:optimization_uta})/(\ref{equ:optimization_ta}) can also be regarded as ranking all operations in the perturbation space and selecting $k$ operations. Therefore, the core question in this paper is how to rank $\Delta G \in \mathbb{T}_{1}(G)$ concerning their relative importance to the attack goal.}

\noindent
\hlt{
\textbf{A. From Design Goals to Evaluation Framework}
}

\noindent
\hlt{
To achieve the above expectation, we propose an evaluation framework to reveal its core requirements.
Our framework composed of three principles, they are:
}

\noindent
\hlt{
(1) \textbf{Benefit decomposition.} 
\hltt{Explaining $B(\cdot)$ is necessary for understanding the three principles.}
When budget $k>1$, perturbations at each step are combined together to function as a whole.
Although final attack results come from collaboration of $k$ different $\Delta G \in \mathbb{T}_{1}(G)$, only one $\Delta G \in \mathbb{T}_{1}(G)$ can be chosen at each perturbation step. 
This fact requires that the attack benefit $B(\cdot)$ should be felicitously defined on $\Delta G \in \mathbb{T}_{1}(G)$ \hltt{from a global view, i.e., distributing $O(\widehat{G};k)$ to each $\Delta G$,} to support the intelligent attacker making perturbation decision.}
\hlt{For example, when using Shapley value $\phi$ \cite{WangRLZ0Z21} as the contribution distribution function, $B(\Delta G) = \phi(\Delta G \big|O(\widehat{G};k))$.
In this paper, an attack method satisfies the \textit{benefit decomposition} principle if it considers all perturbations as a whole to evaluate their attack results, and uses some mechanisms to distribute the overall benefit to the perturbations at each step.}

\noindent
\hlt{(2) \textbf{Baseline graph.} When defining $B(\cdot)$, an implicit ground object is that the \textit{baseline graph} of attack benefit should be fixed on the clean graph $G$, i.e., $B(\cdot)=B(\cdot\big|G)$, and $B(\Delta G) = 0$ where $||\Delta G||=0$.
If the baseline graph changes to other graphs, the measure of attack benefit will deviate from the original meaning (see the example of \textit{GradArgmax} in Appendix B).}

\noindent
\hlt{(3) \textbf{Operation ranking.} 
To achieve stealthy and effective evasion attacks, given specific values of $B(\cdot)$ on $\Delta G \in \mathbb{T}_{1}(G)$, an intelligent attacker should choose the perturbation with maximal attack benefit in each step.
In the above process, the intelligent attacker ranks $\Delta G \in \mathbb{T}_{1}(G)$ according to their attack benefits, and then choose top-$k$ different perturbation operations one-by-one. In this paper, an attack method satisfies the \textit{operation ranking} principle if its perturbation at each step consumes the attack budget by considering the $B(\cdot)$-based ranking of $\Delta G \in \mathbb{T}_{1}(G)$.}

\begin{figure*}[ht!]
  \centering
  \includegraphics[width=\linewidth]{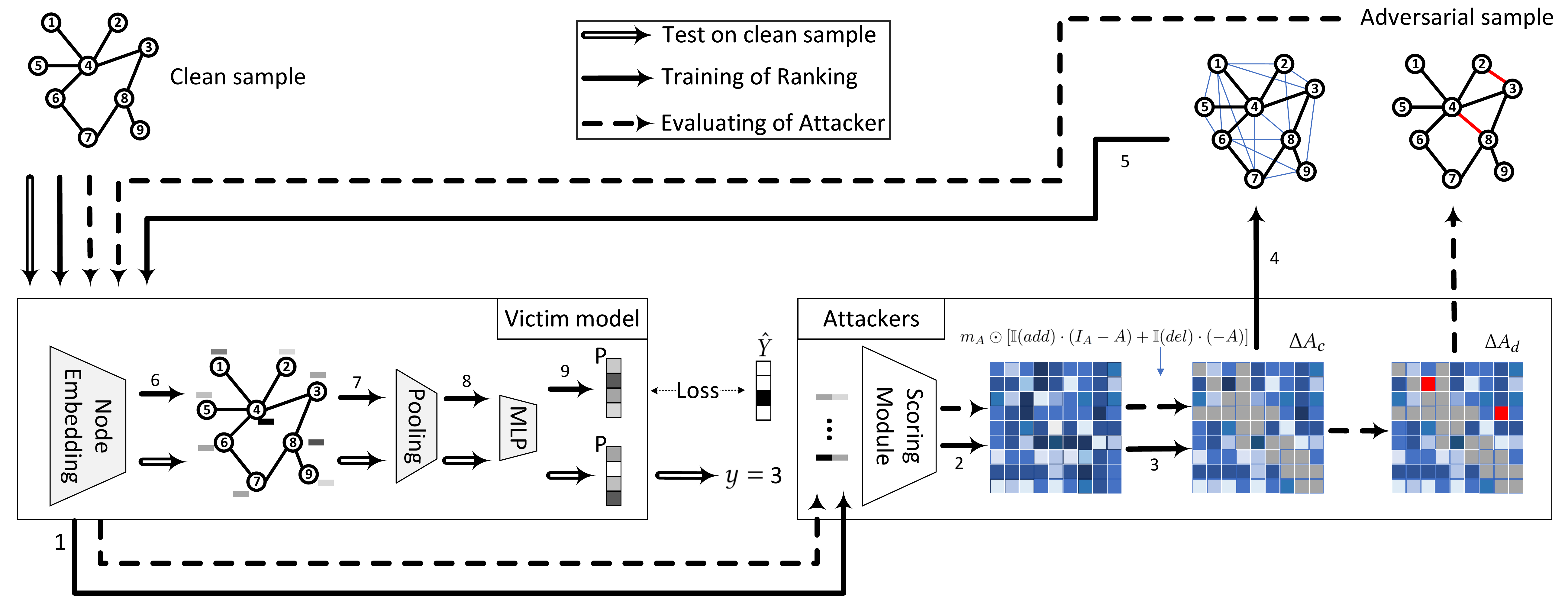}
  \vspace{-18pt}
  \caption{Illustration of the projective ranking method.
The victim model $f_{\theta}$ can make the correct prediction on a clean sample $G$ (shown in hollow arrows). During the scoring module training (shown in the steps 1-9 with solid arrows), the attacker $\mathcal{T}_{f}$ employs the mutual information between the perturbation space $\mathbb{T}(G)$ and the attacker's goal $y_{\widehat{G}}\neq y_{G}$ to measure and rank the importance of all elements in $\mathbb{T}(G)$. The scoring module needs access node embedding and predictive probability distribution of the victim model. In the evaluation phase (shown in dashed arrows), the attacker $\mathcal{T}_{f}$ only needs to access the node embedding to obtain the ranking of the elements in $\mathbb{T}(G)$, and then projects the \textit{operation ranking} to generate adversarial sample $\widehat{G}$ under the specified budget.
}
  \label{fig:frame}
  \vspace{-12pt}
\end{figure*}

\noindent
\hlt{\textbf{B. From Evaluation Framework to Our Method}}

\noindent 
\hlt{In this section, we utilise principles in the evaluation framework and additional requirements to guide the design of our method. Generally, an intelligent attacker always attempts to maximise the profits in attacks (see Appendix A). 
In this paper, we extend the definition of {\itshape transferability} of attack strategies. The {\itshape transferability} means that once attackers learn the attack strategies, they can generate adversarial samples for unseen clean samples under different perturbation budgets. 
However, if attack strategies are learned from fixed budget $k$ (e.g., {\itshape RL-S2V}), these strategy cannot easily satisfy \textit{transferability} since they are designed for specific $k$. Therefore, the evaluation framework and desired transferability practically require an attack method accomplishes $B(\cdot)$-based ranking of $\Delta G \in \mathbb{T}_{1}(G)$ without considering specific $k$ in $B(\cdot)$.}

\hltt{Inspired by the recent research on GNN explainability\cite{ying2019gnnexplainer, DBLP:conf/nips/LuoCXYZC020},} \hlt{we employ all perturbation operation $\Delta G \in \mathbb{T}_{1}(G)$ of the clean graph $G$ (i.e., {\itshape baseline graph}) as a whole to attack target GNNs \hltt{for getting rid of the limitation of specific $k$.} Based on mutual information, the $B(\cdot)$ of each $\Delta G$ is then calculated as their importance (see more in Appendix B) for final attack results. In this way, our method satisfies the \textit{benefit decomposition} principle. To satisfy the \textit{operation ranking} principle, we first rank $\Delta G \in \mathbb{T}_{1}(G)$ according to their importance values (i.e., attack benefit), and then project this ranking into practical attacks with specific $k$.}

\subsubsection{Ranking method}
\hlt{Now, given a clean sample $G$ as the baseline graph, attackers attempt to rank all operation elements in perturbation space $\mathbb{T}(G)$ with respect to their importance.}
We measure the importance of elements in $\mathbb{T}_{1}(G)$ by mutual information ($\mathbf{MI}$) between the modified graph $\widehat{G}$ and the attacker's goal. It can be expressed as:
\begin{equation}
    \max \mathbf{MI}\left( \widehat{Y}, \widehat{G}\right) = H\left(\widehat{Y}\right) - H\left(\widehat{Y} \big| \widehat{G}\right), \label{equ:miyg}
\end{equation}
where $H(\cdot)$ is the entropy function. In untargeted evasion attacks, $\widehat{Y} = \left( \cdots, p_{y_{G}}=0,\cdots \right)$ is the expected prediction distribution of $\widehat{G}$, $y_{G}$ is the original category of $G$. For the clean sample $G$, the first item in (\ref{equ:miyg}) is a constant since the victim model $f_{\theta}$ is fixed and $\widehat{Y}$ is also invariant. So the objective in (\ref{equ:miyg}) is equal to 
\begin{equation}
    \min H\left(\widehat{Y} \big| \widehat{G}\right).
    \label{equ:minhyq}
\end{equation}
To reduce the computational difficulties caused by the discrete graph structure, we apply continuous relaxation on $\Delta G$ and assume it is a graph variable \begin{math}\Delta G \sim \mathcal{T}_{f} (G)\end{math}. Based on Jensen's inequality and $H(\cdot)$ is a concave function, we obtain that
\begin{equation}
\begin{split}
    H\left(\widehat{Y} \big| \widehat{G}\right) 
    &= \sum_{\Delta G \sim \mathcal{T}_{f}(G)}p\left(\Delta G\right)H\left(\widehat{Y}|G+\Delta G\right) \\
    &= E_{\Delta G }\left[ H\left(\widehat{Y}\big|G+\Delta G\right)\right]\\ &\leq H\left(\widehat{Y}\big|G+E\left[\Delta G \right]\right), 
\end{split}
\end{equation}
where $E[\cdot]$ is expectation function. So the objective (\ref{equ:minhyq}) is equal to
\begin{equation}
    \min_{\mathcal{T}_{f} (G)}  H\left(\widehat{Y}\big|G+E_{\mathcal{T}_{f} (G)}\left[\Delta G \right]\right).
\end{equation}
To be more specific, given the victim model $f_{\theta}$, the objective of the untargeted evasion attacks is  
\begin{equation}
    \min_{\mathcal{T}_{f} (G)} -\sum_{y}\mathbb{I}\left(y\neq y_{G}\right) \log p\left(y\big|\widehat{G}\right),
    \label{equ:objfinal}
\end{equation}
where $p\left(\cdot\big|\widehat{G}\right)$ is the predictive probability distribution of $f_{\theta}$ on $\widehat{G}$. The objective of the targeted evasion attacks is
\begin{equation}
\label{equ:objfinal_t}
    \min_{\mathcal{T}_{f} (G)} -\sum_{y}\mathbb{I}\left(y = y_{t}\right) \log p\left(y\big|\widehat{G}\right),
\end{equation}
where $y_{t}$ is the specified target category.

\hlt{Following the attack setting in previous methods \cite{DBLP:conf/icml/DaiLTHWZS18}, we assume that attackers are only able to add edges when generating $\widehat{G}$.}
So we have
\begin{equation}
    \widehat{G} = \mathcal{T}_{f} (G) = \{A+\Delta A, X\},
    \label{equ:tfg}
\end{equation}
where \begin{math}\Delta A= m_{A} \odot \left({I}_{{A}}-{A}\right)\end{math}, the mask $m_{A}$ is obtained by 
\begin{equation}
\begin{split}
    {m_{A}}_{i,j} &= s\left(h_i,h_j\right)\\
    &=softmax\left(MLP(concat(h_i,h_j))\right),
    \label{equ:deltA}
\end{split}
\end{equation}
where $h_i$ is the embedding of node $i$, $s(\cdot,\cdot)$ is the scoring function which measures the importance of elements in $\mathbb{T}_{1}(G)$ with respect to the attacker's expectation. In our method, after concating the embedding of nodes $v_i$ and $v_j$, a Multi Layer Perceptron (MLP) model and the softmax function are employed to obtain the final score. Since the value of $\Delta A$ is a real number rather than a discrete value, we call \begin{math}\widehat{G}\end{math} is generated by adding continuous structure perturbations \begin{math}\Delta G_{c}=\{\Delta A_{c},0\}\end{math}. 

\subsubsection{Perturbation Projection} 

\hlt{In our method, the learned attack strategy is embedded in the ranking of $\Delta G = \{\Delta{A},\Delta{X}\} \in \mathbb{T}_{1}(G)$.}
Due to the discrete structure limitation of graph data, although attackers could obtain the adversarial sample $\widehat{G}$ in (\ref{equ:tfg}) by adding the continuous perturbation $\Delta G_{c}$, it is necessary to bridge the gap between continuous perturbation $\Delta G_{c}$ and discrete modification $\Delta G_{d}$ under limited budgets, 
Inspired by Projected Gradient Descent (PGD) algorithm, we map $\Delta G_{c}$ to $\Delta G_{d}$ by below projection function
\begin{equation}
    \Delta A_{d;i,j} = \mathbb{I}\left(\Delta A_{c;i,j} \in topk\left(\Delta A_{c},k\right)\right),
\end{equation}
where $k$ is the perturbation budget, $topk(\cdot,k)$ is the set of $\Delta A_{c;i,j}$ with top-k values, \hlt{$i$ and $j$ indicate the element at $i$-th row and $j$-th column of matrix.} By this projection, attackers obtain the adversarial sample \begin{math}\widehat{G}=\{A+ \Delta A_{d},X\}\end{math} from clean sample $G$ under specified budget $k$. Figure \ref{fig:frame} illustrates the pipeline of our projective ranking method.


\begin{algorithm}[ht]
\caption{Generation of Adversarial Samples}

\SetKwInput{KwInput}{Input} 
\SetKwInput{KwOutput}{Output} 
\SetKwInput{KwPara}{Parameters} 
\SetKwInput{KwInit}{Initialization} 

\SetKwComment{Comment}{$\triangleright$\ }{}
\SetKwFor{For}{for (}{)}{}
\SetKwFor{While}{while (}{)}{}
\SetKwFor{If}{if (}{)}{}
\SetKwFunction{FMain}{$\mathtt{Main}$}
\SetKwFunction{FRank}{$\mathtt{Ranking}$}
  
\DontPrintSemicolon
  \KwInput{$\mathcal{D} = \{G^{m} = \{A^{m},X^{m}\}\}_{m=1}^M$, classifier $f_{\theta}$}
  \KwOutput{$\widehat{\mathcal{D}} = \{{\widehat{G}}^{m} = \{{\widehat{A}}^{m},X^{m}\}\}_{m=1}^M$}
  \KwPara{$\varphi$ //The trainable parameters of $s$}
  \KwInit{$h = \text{NodeEmbed}\left(\mathcal{D}\big|f_{\theta}\right)$, $ASR_{best} = 0$, $k=1$}

  \SetKwProg{Fn}{Def}{:}{\KwRet}
  \Fn{\FRank{$\mathcal{D}$}}{
        \For{$m = 1;\ m <= M;\ m = m + 1$}{
            $S_{i,j}^{m} = s_{\varphi}(h_{i}^{m},h_{j}^{m})$\;
            $\Delta A_{c}^{m} = \left( {I}_{{A^{m}}}-{A^{m}}\right) \odot S^{m}$\;
        }
        \KwRet $\{\{\Delta A_{c}^{m}\}_{m=1}^{M}\}$\;
  }

 $h = \text{NodeEmbed}\left(\mathcal{D}\big|f_{\theta}\right)$, $ASR_{best} = 0$, $k=1$ \;
  \SetKwProg{Fn}{Function}{:}{\KwRet}
  \Fn{\FMain}{
        \While{\text{not EarlyStop}}{
            // Attacker Training\;
            $\{\Delta A_{c}^{m}\}= \mathtt{Ranking}\left(\mathcal{D}\right)$\;
            $\{\widehat{G}_{c}^{m}\} = \{\{A^{m}+\Delta A_{c}^{m},X^{m}\}\}$\;
            $\min_{\varphi} -\sum_{m=1}^{M}\sum_{y}\mathbb{I}\left(y\neq y_{G}\right) \log p\left(y\big| \widehat{G}_{c}^{m}\right)$\;
            // Attacker Evaluation\;
            $\{\Delta A_{d}^{m}\} = topk(\mathtt{Ranking}\left(\mathcal{D}\right))$\;
            $\widehat{\mathcal{D}} = \{\{A^{m}+\Delta A_{d}^{m},X^{m}\}\}$\;
            // Update the Attack Success Rate\;
            \If{$ASR(\widehat{\mathcal{D}}|f_{\theta}) > ASR_{best}$}{
                $ASR_{best} = ASR_{current}$\;
            }
        }
}
\label{algpipline}
\end{algorithm}

After obtaining the adversarial samples $\widehat{\mathcal{D}} = \{\widehat{G}^{m}\}_{m=1}^M$, the attacker use them to attack the victim model $f_{\theta}$. Algorithm 1 shows how attackers generate adversarial samples from clean samples $\mathcal{D}$
by our projective ranking method. To learn an attack strategy, in line 6, attackers first need to obtain the embedding representation of all nodes $h$ of clean samples in $\mathcal{D}$ from the victim model $f_{\theta}$. In the training phase (lines 9-12), attackers use the scoring function in equation (\ref{equ:deltA}) to obtain structure mask $m_{A}$, then obtain the perturbations on graph structure $\Delta A_{c}$ by limiting $m_{A}$ with the allowed perturbation operation type. Then, in line 11, attackers add perturbations $\Delta A_{c}$ to clean samples to generate a relaxed adversarial sample $\widehat{G}_{c}$. In line 12, attackers use the objective (\ref{equ:objfinal}) to learn an attack strategy embedded in the scoring function. In the evaluation phase (lines 13-18), attackers will obtain the practical adversarial sample $\widehat{G}_{d}$ to evaluate the attack performance of the learned attack strategy. In line 14, attackers project the learned ranking into the practical attack space, in which attackers are only allowed to add edges with the specified budget ($k=1$). Then attackers obtain the final adversarial samples in line 15 and attack the victim model in line 17. Attackers will repeat the process (lines 9-18) until the projective ranking method learned a high attack strategy. By replacing the objective in line 12 with objective (\ref{equ:objfinal_t}), attackers could obtain the adversarial samples under the setting of targeted attacks. 

\begin{table}[t]
\vspace{-8pt}
\centering
\caption{Consistency with the Principles of the Evaluation Framework}
\vspace{-5pt}
\label{tab:consistency}
\begin{threeparttable}
\begin{tabular}{@{}cccc@{}}
\toprule
\multirow{2}{*}{method} &
  \multirow{2}{*}{\begin{tabular}[c]{@{}c@{}}benefit\\ decomposition\end{tabular}} &
  \multirow{2}{*}{\begin{tabular}[c]{@{}c@{}}baseline\\ graph\end{tabular}} &
  \multirow{2}{*}{\begin{tabular}[c]{@{}c@{}}operation\\ ranking\end{tabular}} \\
  &  &  &  \\ \midrule
Random  & \emptycirc[0.8ex] & \fullcirc[0.8ex] & \emptycirc[0.8ex] \\
GradArgmax & \emptycirc[0.8ex] & \emptycirc[0.8ex] & \emptycirc[0.8ex]  \\
RL-S2V  & \fullcirc[0.8ex] & \fullcirc[0.8ex] & \emptycirc[0.8ex] \\
Our  & \fullcirc[0.8ex] & \fullcirc[0.8ex] & \fullcirc[0.8ex]\\
\bottomrule
\end{tabular}
\begin{tablenotes}
\item \emptycirc[0.8ex] means neglect, \fullcirc[0.8ex] means consideration.
\end{tablenotes}
\end{threeparttable}
\vspace{-10pt}
\end{table}

\section{Method Analysis under Our Evaluation Framework}
\label{sec:frameworks_evaluation}

In this section, we use the three principles in Section \ref{sec:evaluationframe} to make an evaluation on our method and three typical attack methods: \textit{RandomSampling}, \textit{GradArgmax} and \textit{RL-S2V}. The results are summarized in the Table \ref{tab:consistency}. In these methods, \textit{RandomSampling} only satisfies the {\itshape baseline graph} principle, our method satisfies the \textit{attack benefit}, \textit{operation ranking} and {\itshape baseline graph} simultaneously. The other methods are partially consistent with the evaluation framework. The reasons are as follows.

\noindent
\textbf{(1) GradArgmax.} \hlt{Similar to the use of gradient descent for training in neural networks, {\itshape GradArgmax} employs the gradient of $f_{\theta}$ on all {\itshape operation elements} $\Delta G_{cur}\in \mathbb{T}_{1}(G_{cur})$ to choose perturbation operation, where $G_{cur}$ is the graph waiting to be modified at the current perturbation step.} Hence, {\itshape GradArgmax} neglects the {\itshape baseline graph} principle. To use up all perturbation budget $k$, attackers need to calculate the gradient information at each perturbation step. Then it will use the real-time local gradient as the measure of attack benefits. Actually, the gradient information is a measure of modification sensitivity (see Appendix B), so {\itshape GradArgmax} does not satisfy the \textit{attack benefit} principle. \hlt{To generate adversarial samples, {\itshape GradArgmax} focuses on integrating all perturbation graph $\Delta G_{cur}\in \mathbb{T}_{1}(G_{cur})$, which owns the biggest local benefit at each perturbation step.} Although the operations in \textit{GradArgmax} have an order, it does not satisfy the  {\itshape operation ranking} principle since the local gradient information is not a suitable measure for attack benefits.

\noindent
\textbf{(2) RL-S2V.} In the generation of adversarial samples, {\itshape RL-S2V} uses the attack benefit of $\Delta G\in \mathbb{T}_{k}(G)$ to decide the perturbation operation $\Delta G\in \mathbb{T}_{1}(G)$ of each step in consuming perturbation budget $k$. The reward function in \textit{RL-S2V} is designed by considering the outputs of the victim model $f_{\theta}$ on $\widehat{G}$ and $G$, so it satisfies the \textit{baseline graph} and \textit{attack benefit} requirements. However, it emphasizes using $\Delta G\in \mathbb{T}_{k}(G)$ as a whole perturbation to attack the victim model $f_{\theta}$ with specified budget $k$, while cares less about the order of the element $\Delta G\in \mathbb{T}_{1}(G)$ that composes the perturbation space $\mathbb{T}(G)$. Hence, the {\itshape RL-S2V} attack method neglects the {\itshape attack ranking}. 

Furthermore, we analyze the difference between the measures of attack benefits (sensitivity, long-term benefit, importance) in the Appendix B. We also show the weakness of \textit{GradArgmax} concerning the attack benefits from three different aspects and empirically demonstrate them in our experimental results.



\section{Experiments}
\label{sec:experiments}

\subsection{Datasets and Baselines}
We employ several real-world datasets ENZYMES, Mutagenicity, PC-3, NCI109, NCI-H23H \cite{Morris+2020} and one synthesized dataset BA-2Motifs \cite{DBLP:conf/nips/LuoCXYZC020} for graph classification. For the NCI-H23H dataset, we only choose the graphs whose node size is less than 50. Table \ref{tab:datasta} shows some basic statistics of these datasets.

To evaluate the performance of our method, we select \textit{RandomSampling}, \textit{GradArgmax}, and \textit{RL-S2V} \cite{DBLP:conf/icml/DaiLTHWZS18} as baselines. In the \textit{RandomSampling} method, we performed ten times with different seeds and then averaged these accuracy results. In our experiments, the attack performance is measured by the accuracy of the victim model, in which a low accuracy number indicates a high attack performance.

\begin{table}[t!]
    \centering
    \caption{Datasets Statistics}
    \vspace{-5pt}
    \label{tab:datasta}
    \begin{tabular}{lccc}
        \toprule
        Dataset & $\#$ of Graphs & $\#$ of Classes & Avg. $\#$ of Edges \\
        \midrule
        ENZYMES & 600 & 6 & 62.14\\
        Mutagenicity& 4337 & 2 & 30.77 \\
        PC-3  & 27509 & 2 & 28.49 \\
        NCI109  & 4127 & 2 & 32.13\\
        NCI-H23H & 26838 & 2 & 37.27\\
        BA-2Motifs& 1,000 & 2 & 25.48\\
        \bottomrule
    \end{tabular}
    \vspace{-12pt}
\end{table}

\begin{table*}[h!]
\centering
\caption{Classification Accuracy of the Victim GAT Model (\%). The upper half shows the attack results of different attack methods on seen samples, and the lower half shows the attack results on unseen samples. $k$ is the perturbation budget.}
\vspace{-5pt}
\label{tab:attack_gat}
\begin{threeparttable}
\begin{tabular}{cccccccccccccccc}
\toprule
Dataset & \multicolumn{3}{c}{ENZYMES} & \multicolumn{3}{c}{Mutagenicity} & \multicolumn{3}{c}{PC-3} & \multicolumn{3}{c}{NCI109} & \multicolumn{3}{c}{NCI-H23H} \\
\midrule
k & 1 & 2 & 3 & 1 & 2 & 3 & 1 & 2 & 3 & 1 & 2 & 3 & 1 & 2 & 3 \\
\midrule
Clean & 65.83 & 65.83 & 65.83 & 73.63 & 73.63 & 73.63 & 51.75 & 51.75 & 51.75 & 76.50 & 76.50 & 76.50 & 71.25 & 71.25 & 71.25 \\
\midrule
Rand\tnote{*} & 60.04 & 51.71 & 45.79 & 73.39 & 70.86 & 69.08 & \underline{\textit{51.85}} & 51.74 & 51.55 & 74.80 & 71.25 & 68.50 & 70.80 & 66.48 & 63.80 \\
\midrule
Grad\tnote{*} & \underline{\textit{65.83}} & \underline{\textit{65.83}} & \underline{\textit{65.83}} & \underline{\textit{73.63}} & \underline{\textit{73.63}} & \underline{\textit{73.63}} & \underline{\textit{51.75}} & \underline{\textit{51.75}} & \underline{\textit{51.75}} & \underline{\textit{76.50}} & \underline{\textit{76.50}} & \underline{\textit{76.50}} & \underline{\textit{71.25}}  & \underline{\textit{71.25}} & \underline{\textit{71.25}} \\
\midrule
RL-S2V & 53.75 & \underline{\textit{52.08}} & \underline{\textit{51.46}} & \textbf{69.38} & \textbf{68.25} & 68.50 & \textbf{51.13} & \textbf{51.13} & \textbf{50.75} & 69.88 & 68.13 & 67.25 & \textbf{62.38} & 62.50  & 60.50  \\
\midrule
\textbf{Ours} & \textbf{48.75} & \textbf{39.17} & \textbf{33.13} & 71.88 & 69.63 & \textbf{68.25} & 51.50 & 51.25 & 51.00 & \textbf{67.75} & \textbf{63.38} & \textbf{60.13} & 62.87 & \textbf{55.13} & \textbf{49.25} \\
\bottomrule
\toprule
\multicolumn{16}{c}{Transfer Attack of the Learned Strategy on Unseen Samples} \\
\midrule
Dataset & \multicolumn{3}{c}{ENZYMES} & \multicolumn{3}{c}{Mutagenicity} & \multicolumn{3}{c}{PC-3} & \multicolumn{3}{c}{NCI109} & \multicolumn{3}{c}{NCI-H23H} \\
\midrule
k & 1 & 2 & 3 & 1 & 2 & 3 & 1 & 2 & 3 & 1 & 2 & 3 & 1 & 2 & 3 \\
\midrule
Clean & 64.17 & 64.17 & 64.17 & 75.74 & 75.74 & 75.74 & 95.60 & 95.60 & 95.60 & 78.18 & 78.18 & 78.18 & 84.54 & 84.54  & 84.54  \\
\midrule
Rand & 63.08 & 58.08 & 53.33 & 74.78 & 72.59 & 71.02 & 95.60 & 95.58 & 95.56 & 76.69 & 73.14 & 70.42 & 82.99 & 80.05 & 77.40 \\
\midrule
Grad & - & - & - & - & - & - & - & - & - & - & - & - & - & - & - \\
\midrule
RL-S2V & \textbf{59.17} & \underline{\textit{59.17}} & \underline{\textit{58.33}} & \textbf{73.42} & \underline{\textit{73.17}} & \underline{\textit{72.43}} & \textbf{95.59} & \underline{\textit{95.59}} & \underline{\textit{95.62}} & 74.99 & \underline{\textit{73.82}} & \underline{\textit{72.26}} & 81.75 & \underline{\textit{81.12}} & 76.72 \\
\midrule
\textbf{Ours} & 63.33 & \textbf{52.50} & \textbf{45.83} & 74.72 & \textbf{71.67} & \textbf{69.52} & 95.60 & \textbf{95.57} & \textbf{95.54} & \textbf{71.90} & \textbf{67.18} & \textbf{63.81} & \textbf{77.92} & \textbf{72.34} & \textbf{66.56}\\
\bottomrule
\end{tabular}

\begin{tablenotes}
    \footnotesize \item[*]Rand is the \textit{RandomSampling} attack method, and Grad is the \textit{GradArgmax} attack method.
\end{tablenotes}
\end{threeparttable}
\vspace{-5pt}
\end{table*}

\begin{table*}
\centering
\caption{Classification Accuracy of the Victim GCN Model (\%). The upper half shows the attack results of different attack methods on seen samples, and the lower half shows the attack results on unseen samples. $k$ is the perturbation budget.}
\vspace{-5pt}
\label{tab:attack_gcn}
\begin{threeparttable}
\begin{tabular}{cccccccccccccccc}
\toprule
Dataset & \multicolumn{3}{c}{ENZYMES} & \multicolumn{3}{c}{Mutagenicity} & \multicolumn{3}{c}{PC-3} & \multicolumn{3}{c}{NCI109} & \multicolumn{3}{c}{NCI-H23H} \\
\midrule
k & 1 & 2 & 3 & 1 & 2 & 3 & 1 & 2 & 3 & 1 & 2 & 3 & 1 & 2 & 3 \\
\midrule
Clean & 69.17 & 69.17 & 69.17 & 85.25 & 85.25 & 85.25 & 67.13 & 67.13 & 67.13 & 76.50 & 76.50 & 76.50 & 64.75 & 64.75 & 64.75 \\
\midrule
Rand & 64.81 & 56.33 & 51.23 & 80.80 & 75.56 & 70.75 & 64.60 & 60.10 & 57.64 & 73.83 & 67.60 & 63.29 & 60.06 & 55.26 & 53.51 \\
\midrule
Grad & 64.79 & \underline{\textit{61.46}} & \underline{\textit{58.13}} & 68.13 & \textbf{54.50} & \textbf{45.50} & \underline{\textit{66.13}} & \underline{\textit{65.38}} & \underline{\textit{65.25}} & \underline{\textit{77.00}} & \underline{\textit{74.00}} & \underline{\textit{71.00}} & 56.13 & 53.25 & 52.63 \\
\midrule
RL-S2V & 58.54 & 56.04 & \underline{\textit{55.42}} & 71.88 & 69.50 & 63.75 & 57.88 & 55.88 & 55.00 & \textbf{66.00} & 65.00 & \underline{\textit{64.13}} & 54.38 & 52.88 & 52.00 \\
\midrule
\textbf{Ours} & \textbf{52.50} & \textbf{44.79} & \textbf{42.92} & \textbf{67.63} & 59.13 & 53.63 & \textbf{56.13} & \textbf{52.13} & \textbf{49.62} & 68.38 & \textbf{59.00} & \textbf{56.75} & \textbf{53.13} & \textbf{50.25} & \textbf{50.13} \\
\bottomrule
\toprule
\multicolumn{16}{c}{Transfer Attack of the Learned Strategy on Unseen Samples} \\
\midrule
Dataset & \multicolumn{3}{c}{ENZYMES} & \multicolumn{3}{c}{Mutagenicity} & \multicolumn{3}{c}{PC-3} & \multicolumn{3}{c}{NCI109} & \multicolumn{3}{c}{NCI-H23H} \\
\midrule
k & 1 & 2 & 3 & 1 & 2 & 3 & 1 & 2 & 3 & 1 & 2 & 3 & 1 & 2 & 3 \\
\midrule
Clean & 69.17 & 69.17 & 69.17 & 84.93 & 84.93 & 84.93 & 96.37 & 96.37 & 96.37 & 78.45 & 78.45 & 78.45 & 98.41 & 98.41 & 98.41 \\
\midrule
Rand & 66.83 & 59.67 & 54.25 & 81.01 & 75.62 & 70.81 & 95.69 & 94.57 & 93.39 & 76.17 & 70.23 & 65.70 & 98.34 & 98.14 & 97.99 \\
\midrule
Grad & - & - & - & - & - & - & - & - & - & - & - & - & - & - & - \\
\midrule
RL-S2V & 63.33 & \underline{\textit{62.50}} & \underline{\textit{61.67}} & \textbf{74.39} & 74.07 & 66.95 & \underline{\textit{95.75}} & \underline{\textit{95.58}} & \underline{\textit{94.91}} & \textbf{74.03} & \underline{\textit{72.47}} & \underline{\textit{69.40}} & 98.30 & \textbf{97.69} & 97.63 \\
\midrule
\textbf{Ours} & \textbf{60.00} & \textbf{51.67} & \textbf{44.17} & 74.44 & \textbf{63.84} & \textbf{57.45} & \textbf{95.11} & \textbf{93.14} & \textbf{91.74} & 74.12 & \textbf{67.33} & \textbf{63.63} & \textbf{98.23} & 97.82 & \textbf{97.55}
 \\
\bottomrule
\end{tabular}
\end{threeparttable}
\vspace{-8pt}
\end{table*}

\vspace{-5pt}
\subsection{Experimental Setup}
In the victim models, we randomly split each dataset into training data ($80\%$), validation data ($10\%$), and test data ($10\%$) to train the victim GAT \cite{velivckovic2017graph} or  GCN \cite{DBLP:conf/iclr/KipfW17} models. The victim models have three hidden layers, with 20 as the output feature dimension. We concatenate the $maxpooling$ and $sumpooling$ results of the final embedding of nodes to obtain the expression of the whole graph, then use a fully connected layer to predict the category of the graph. In our projective ranking method, we use a 2-layer MLP to serve as the scoring module. 

To evaluate the {\itshape transferability} of our method, we directly project the learned strategy under $k=1$ to the generation of adversarial samples under $k=2,3$. The other results are obtained by running corresponding attack methods under a specified budget, respectively.

\vspace{-5pt}
\subsection{Attack Performance Comparison}
\label{exp:transk_ut}
\subsubsection{Untargeted Attacks}
The upper half of Table \ref{tab:attack_gat} and \ref{tab:attack_gcn} show the accuracy of the victim model $f_{\theta}$ on both clean and adversarial samples generated by different attack methods. The bold numbers indicate the best attack results under the same setting. 

In Table \ref{tab:attack_gat} and \ref{tab:attack_gcn}, our method achieves the best or competitive attack performance on all datasets with $k$=1, indicating the projective ranking method owns {\itshape powerful} attack performance. When $k$=2 or 3, we use the learned attack strategy under $k$=1 to generate adversarial samples for the changed perturbation budget. The results show our method also obtain the best or competitive attack effect, which shows that the projective ranking method has learned an effective {\itshape transferable} attack strategy under budget $k$=1. 

\begin{figure}[ht]
  \centering
  \includegraphics[width=\linewidth]{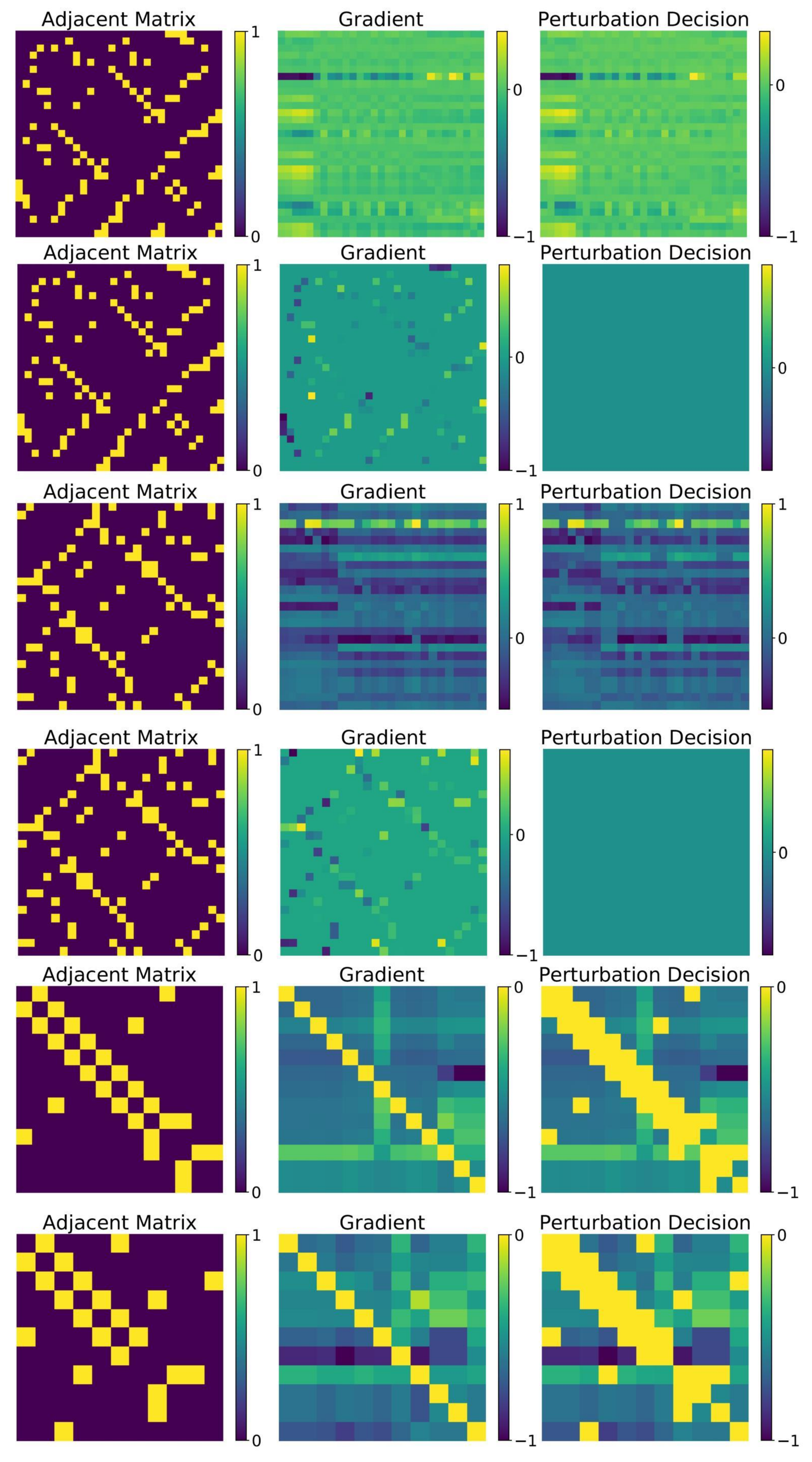}
  \vspace{-18pt}
  \caption{Visualization of the graph structure, gradient information and perturbation decision in \textit{GradArgmax}. The 1st and 2nd rows of images come from the same sample under the GCN and GAT models, respectively. Compared with the victim GCN model, the nonzero gradients from the GAT model almost locate at the same position as the graph structure. The zero-gradient at the location of all possible edge adding operations invalidates the \textit{GradArgmax} attack method on the victim GAT model. The 3rd row of images is a sample based on the GCN model. Although this sample has nonzero perturbation decision values on the edge adding positions, the decision value is negative. Adding edges on this kind of sample is also unavailable since adding edges on any position causes negative attack results. \hlt{In this figure, we have scaled the values in the middle and right columns to [-1,1] for better visualisation.}}
  \label{fig:gradviz}
  \vspace{-12pt}
\end{figure}

\textit{GradArgmax} generates the adversarial examples based on the gradient of the victim model and makes a greedy choice in each step. The attack results of \textit{GradArgmax} in the Table \ref{tab:attack_gat} and \ref{tab:attack_gcn} empirically prove our discussion about sensitivity in Appendix B. Firstly, the perturbation generated by \textit{GradArgmax} is sub-optimal since the gradient is the sensitivity measure of the victim model. Although the \textit{GradArgmax} achieves effective attack performance on the victim GCN model, its ability on the victim GAT model is limited (see Figure \ref{fig:gradviz}). Secondly, the result on the NCI109 dataset with $k$=1 in Table \ref{tab:attack_gcn} shows that \textit{GradArgmax} generates samples with negative attack performance. This indicates that its attack performance is limited by the non-linear nature of the GNN model in some situations. Finally, the attack performance of \textit{GradArgmax} is worse than the \textit{RandomSampling} on ENZYMES, PC-3 and NCI109 datasets when $k$=2 and 3, on which \textit{GradArgmax} shows weak marginal attack performance concerning the budget. Moreover, the \textit{GradArgmax} does not learn any attack strategy in the attack process. The above results indicate that \textit{GradArgmax} is not desirable for attackers in some situations since it misses the \textit{baseline graph} and \textit{attack benefit} principles in evasion attacks.

Compare with using the ``local" gradient in \textit{GradArgmax}, \textit{RL-S2V} and our method achieve stable attack performance on both GAT and GCN models since both of them consider the \textit{importance} of perturbation operation with ``global" criteria. We also use \textit{RandomSampling} as a baseline to evaluate other attack methods. Table \ref{tab:attack_gat} and \ref{tab:attack_gcn} show our method always achieve better attack performance than \textit{RandomSampling}. The $\underline{italic}$ shows unexpected results in the aspect of attack ability, where the performance of \textit{RL-S2V} is worse than \textit{RandomSampling} on ENZYMES ($k$=2,3) in Table \ref{tab:attack_gat} and ENZYMES ($k$=3), NCI109 ($k$=3) in Table \ref{tab:attack_gcn}. Moreover,  attackers always expect to obtain better attack performance when giving more budgets. If the budget increases, a well-designed attack method should not decrease its performance. However, like the result in \cite{ma2019attacking}, we also observed that the attack performance of \textit{RL-S2V} degrades when increasing the perturbation budget on Mutagenicity ($k$=3) in Table \ref{tab:attack_gat}. This may be due to the less effectiveness of the Q-learning method in \textit{RL-S2V}, especially for the Markov decision process with a long horizon.

\begin{table}[ht]
\centering
\caption{\hlt{Attack Success Rate (\%) of Targeted Attacks on ENZYMES.}}
\vspace{-5pt}
\label{tab:targeted_attack}
\begin{tabular}{ccccccc}
\toprule
Model  & \multicolumn{3}{c}{GCN} & \multicolumn{3}{c}{GAT} \\ \midrule
\textit{k}      & 1     & 2      & 3      & 1      & 2      & 3     \\ \midrule
Rand   & 2.03  & 4.45   & 7.53   & 2.45   & 5.60   & 7.90  \\ \midrule
Grad   & 4.50  & 10.75  & 15.75  & \underline{\textit{0.00}}   & \underline{\textit{0.00}}   & \underline{\textit{0.00}}  \\ \midrule
RL-S2V & 7.25  & 9.50   & 9.50   & 5.50   & 6.25   & \underline{\textit{5.25}}  \\ \midrule
\textbf{Ours} & \textbf{10.75} & \textbf{16.00} & \textbf{19.25} & \textbf{9.50} & \textbf{14.75} & \textbf{17.50} \\ \bottomrule \toprule
\multicolumn{7}{c}{Transfer Attack of   the Learned Strategy on Unseen Samples}                                    \\ \midrule
\textit{k}      & 1     & 2      & 3      & 1      & 2      & 3     \\ \midrule
Rand   & 1.40  & 2.80   & 5.00   & 2.10   & 4.20   & 4.60  \\ \midrule
Grad   & -     & -      & -      & -      & -      & -     \\ \midrule
RL-S2V & 3.00  & 6.00   & \underline{\textit{4.00}}   & 3.00   & \underline{\textit{4.00}}   & \underline{\textit{1.00}}  \\ \midrule
\textbf{Ours} & \textbf{7.00}  & \textbf{9.00}  & \textbf{14.00} & \textbf{8.00} & \textbf{15.00} & \textbf{18.00} \\ \bottomrule
\end{tabular}
\vspace{-12pt} 
\end{table}

\subsubsection{Targeted Attacks}
\hlt{We also conduct targeted attacks on the ENZYME dataset. Without loss of generality, these methods attempt to attack GNN models so that they classify graphs from other classes (i.e., 1-5) as class 0. The upper half of Table \ref{tab:targeted_attack} shows the attack success rate of different attack methods, and our method far outperforms other baselines in attack performance. Moreover, in targeted attacks, we also found the same weaknesses of \textit{GradArgmax} and \textit{RL-S2V} that they have in untargeted attacks (see $\underline{italic}$ results).}

\noindent
\hlt{\textit{Remarks.} \hltt{Although the attention mechanism in GAT models introduces more parameters, previous research \cite{LiuAQCFYH21} in imbalanced learning shows that GAT's performance is lower than GCN's because the minority class does not have enough data to train a competent model.
} When training the GAT model on the NCI-H23H dataset, we employ under-sampling on the category with most samples to alleviate the unfavourable effect caused by unbalanced distribution. In the training data, the \# of positive samples: \# of negative samples = 1:2.}

\hlt{All results in this section are achieved by directly applying attack methods on the vanilla GNN models. Recently, the randomized smoothing method \cite{CohenRK19, JiaCWG20} has been widely employed to improve the robustness of AI models. In this paper, we follow the setting (noise parameter $\beta = 0.9/0.7$, sampling number $d=10000$) of recent research on certified robustness of GNNs \cite{WangJCG21} to evaluate the attack performance of our method. 
Experimental results show randomized smoothing \cite{WangJCG21} is an effective defence method. However, for GNN models in this paper, randomized smoothing is not applicable because it significantly degrades the performance (i.e., accuracy) of GNNs. 
For example, as shown in Table \ref{tab:defence}, although our method almost cannot attack the GCN models on ENZYMES/Mutagenicity after introducing randomized smoothing, the accuracy of GCN equipped with the defence method ($\beta=0.9$) on clean graphs is only 22.29\%/53.00\%, which is dramatically lower than that of vanilla GCN models suffering attacks with $k=1$ (i.e., 52.50\%/67.63\%).}
\begin{table}[ht!]
\centering
\caption{\hlt{Accuracy (\%) Comparison of the Victim Models without/with Defence.}}
\vspace{-5pt}
\label{tab:defence}
\begin{threeparttable}
\begin{tabular}{@{}cccccccc@{}}
\toprule
\multicolumn{2}{c}{Dataset}                                                    & \multicolumn{3}{c}{ENZYMES} & \multicolumn{3}{c}{Mutagenicity} \\ \midrule
\multicolumn{2}{c}{k}        & 1     & 2     & 3     & 1     & 2     & 3     \\ \midrule
\multirow{2}{*}{W/o\tnote{*}} & Clean & 69.17 & 69.17 & 69.17 & 85.25 & 85.25 & 85.25 \\ \cmidrule(l){2-8} 
                     & Ours  & 52.50  & 44.79 & 42.92 & 67.63 & 59.13 & 53.63 \\ \midrule
\multirow{2}{*}{\begin{tabular}[c]{@{}c@{}}W \tnote{*} \\ $\beta=0.9$\end{tabular}} & Clean & 22.29   & 22.29   & 22.29   & 53.00     & 53.00     & 53.00    \\ \cmidrule(l){2-8} 
                     & Ours  & 22.29 & 22.29 & 22.29 & 52.37 & 52.13 & 52.00 \\ \midrule
\multirow{2}{*}{\begin{tabular}[c]{@{}c@{}}W\\ $\beta=0.7$\end{tabular}}  & Clean & 17.50   & 17.50   & 17.50   & 49.13     & 49.13     & 49.13    \\ \cmidrule(l){2-8} 
                     & Ours  & 17.50 & 17.50 & 17.50 & 49.25 & 49.00 & 49.00 \\ \bottomrule
\end{tabular}
\begin{tablenotes}
    \footnotesize \item[*] W/o indicates results on vanilla GCN models, and W indicates results on GCN models equipped with randomize smoothing. $\beta$ presents the noise parameter in the randomized smoothing.
\end{tablenotes}
\end{threeparttable}
\vspace{-12pt} 
\end{table}

\subsection{Transferability on Unseen Samples}
\label{exp:transunseen}
To reduce attackers' effort to generate adversarial samples from unseen clean samples, they generally expect a transferable attack strategy. In this section, we use the attack strategies obtained in Section \ref{exp:transk_ut} to generate adversarial samples for unseen samples and then employ them to attack the victim models. 

\subsubsection{Untargeted Attacks}
The lower half of Tables \ref{tab:attack_gat} and \ref{tab:attack_gcn} show the classification accuracy of $f_{\theta}$ on both clean and adversarial samples. We consider the \textit{RL-S2V} and our projective ranking method since only they can learn transferable attack strategies. The bold results indicate the best method under the same setting, and \underline{\textit{italic}} show unexpected results in the aspect of transferable attack performance. Firstly, the results in Tables \ref{tab:attack_gat} and \ref{tab:attack_gcn} show both \textit{RL-S2V} and our method could use the learned attack strategy to generate effective adversarial samples under perturbation budget $k$=1 on all datasets. It indicates that the learned strategy can be transferred to attack unseen samples. Secondly, the results under $k$=2, 3 further demonstrate that the projective ranking method owns transferable attack performance on unseen samples when the budget changes. Especially, our method obtains the best attack performance on some datasets when $k$=2 or 3 even if it does not own the best performance when $k$=1. Finally, compared to \textit{RandomSampling}, the transferability of the attack strategy of \textit{RL-S2V} is limited on most datasets when $k$=2 or 3, maybe the \textit{RL-S2V} did not learn a powerful attack policy when $k>1$. These results reveal that the victim models make consistent mistakes at both the sample and the budget levels. The attackers can utilize this fact to learn the attack strategy only under a specific budget with limited resources and then use the attack strategies to attack the target classifiers when the attack scenario changes.

\begin{figure*}[ht!]
  \centering
  \includegraphics[scale=0.69]{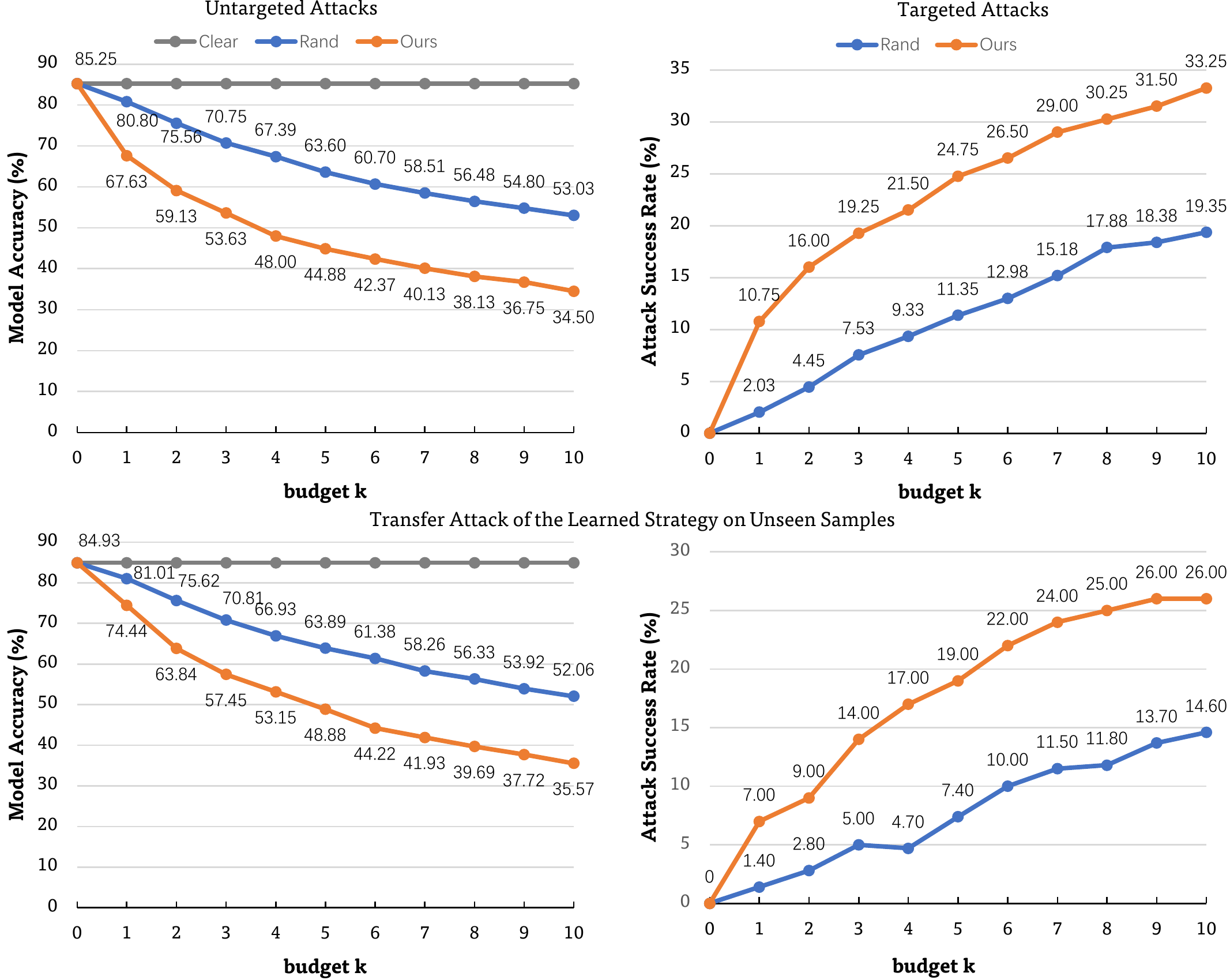} 
  \vspace{-12pt}
  \caption{\hlt{Attack performance of our methods under different perturbation budgets. The figures in the left/right column show untargeted/targeted attack results (i.e., model accuracy, or attack success rate) on Mutagenicity/ENZYMES dataset. The figures in the first/second row show attack results on seen/unseen samples.}}
  \label{fig:attacks_different_k}
  \vspace{-12pt}
\end{figure*}

\begin{figure}[ht!]
  \centering
  \includegraphics[width=\linewidth]{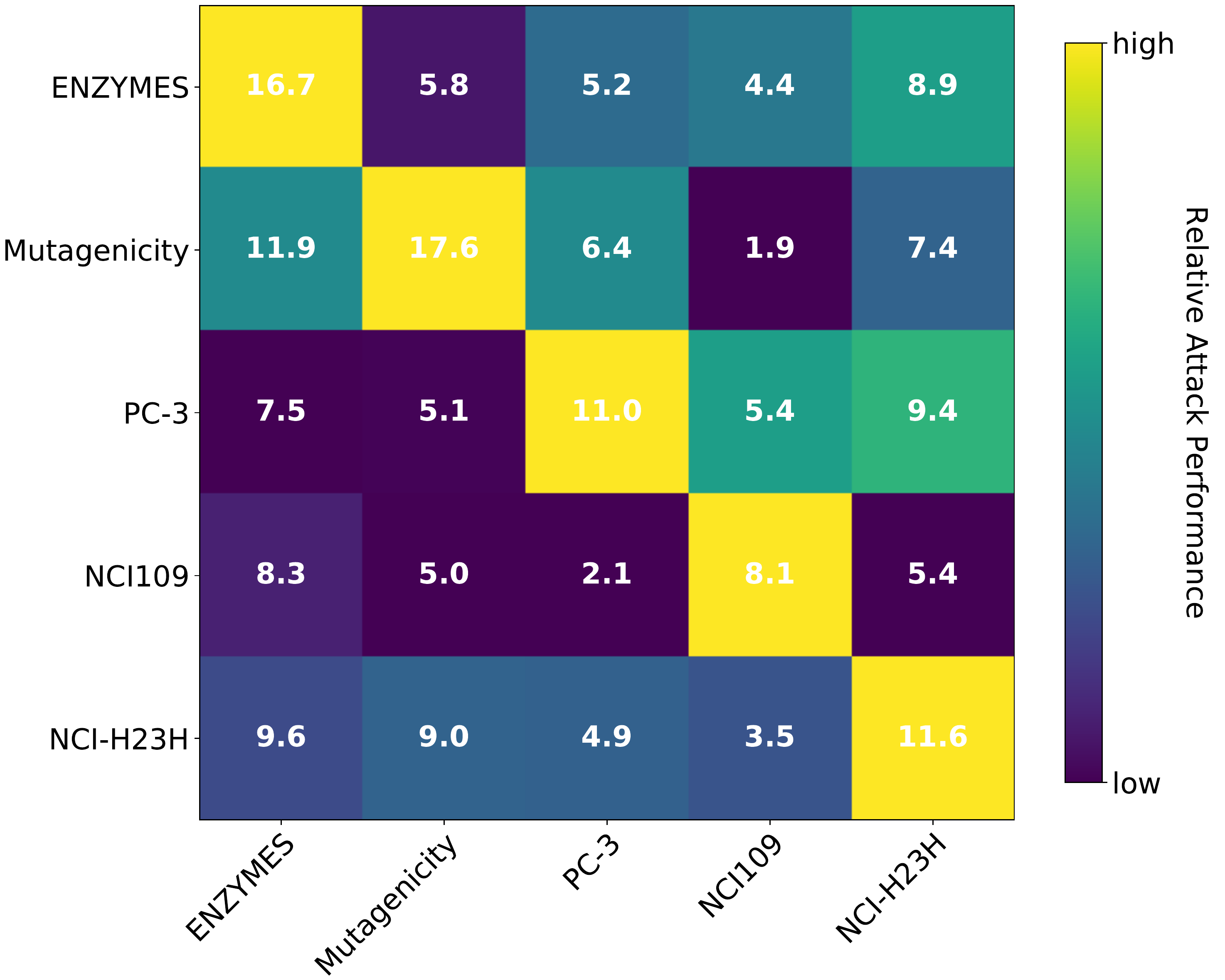}
  \vspace{-18pt}
  \caption{Visualization of transferable attack performance (k=1). The row is the dataset on which the source attacker is trained, and the column is the dataset of the target GCN model. These numbers indicate how much the classification accuracy of the target classifiers have decreased, and a higher value means a stronger attack capability. The color indicates the relative attack performance.}
  \label{fig:crossdomaingcn}
  \vspace{-12pt}
\end{figure}

\begin{table*}[ht!]
\centering
\caption{Classification Accuracy of the Victim GAT/GCN Model (\%).}
\vspace{-5pt}
\label{tab:attack_gatgcn}
\begin{threeparttable}
\begin{tabular}{cccccccccccccccc}
\toprule
Dataset & \multicolumn{3}{c}{ENZYMES} & \multicolumn{3}{c}{Mutagenicity} & \multicolumn{3}{c}{PC-3} & \multicolumn{3}{c}{NCI109} & \multicolumn{3}{c}{NCI-H23H} \\
\midrule
k & 1 & 2 & 3 & 1 & 2 & 3 & 1 & 2 & 3 & 1 & 2 & 3 & 1 & 2 & 3 \\
\midrule
Clean & 65.83 & 65.83 & 65.83 & 73.63 & 73.63 & 73.63 & 51.75 & 51.75 & 51.75 & 76.50 & 76.50 & 76.50 & 71.25 & 71.25 & 71.25 \\
\midrule
Rand & 60.04 & 51.71 & 45.79 & 73.39 & 70.86 & 69.08 & \underline{\textit{51.85}} & 51.74 & 51.55 & \textbf{74.80} & 71.25 & 68.50 & 70.80 & 66.48 & 63.80  \\
\midrule
\textbf{GCN-GAT}\tnote{*} & \textbf{57.50} & \textbf{49.38} & \textbf{44.17} & \textbf{71.25} & \textbf{65.88} & \textbf{62.88} & \textbf{51.50} & \textbf{51.38} & \textbf{51.25} & 75.25 & \textbf{70.50} & \textbf{67.13} & \textbf{69.38} & \textbf{66.38} & \textbf{63.75} \\
\bottomrule
\toprule
Dataset & \multicolumn{3}{c}{ENZYMES} & \multicolumn{3}{c}{Mutagenicity} & \multicolumn{3}{c}{PC-3} & \multicolumn{3}{c}{NCI109} & \multicolumn{3}{c}{NCI-H23H} \\
\midrule
k & 1 & 2 & 3 & 1 & 2 & 3 & 1 & 2 & 3 & 1 & 2 & 3 & 1 & 2 & 3 \\
\midrule
Clean & 69.17 & 69.17 & 69.17 & 85.25 & 85.25 & 85.25 & 67.13 & 67.13 & 67.13 & 76.50 & 76.50 & 76.50 & 64.75 & 64.75 & 64.75 \\
\midrule
Rand & 64.81 & 56.33 & 51.23 & 80.80 & 75.56 & 70.75 & 64.60 & 60.10 & 57.64 & 73.83 & 67.60 & 63.29 & 60.06 & 55.26 & 53.51 \\
\midrule
\textbf{GAT-GCN} & \textbf{56.04} & \textbf{43.96} & \textbf{36.46} & \textbf{80.63} & \textbf{74.13} & \textbf{68.63} & \textbf{62.50} & \textbf{58.75} & \textbf{55.50} & \textbf{68.00} & \textbf{62.00} & \textbf{57.75} & \textbf{56.75} & \textbf{51.88} & \textbf{50.75}\\
\bottomrule
\end{tabular}
\begin{tablenotes}
    \footnotesize \item[*] This means we use the adversarial sample generated from the GCN model to attack the victim GAT model.
\end{tablenotes}
\end{threeparttable}
\vspace{-12pt}
\end{table*}

\subsubsection{Targeted Attacks}
\hlt{The lower half of Table \ref{tab:targeted_attack} shows attack success rate of attack methods on unseen samples. Attack results show our method observably outperforms \textit{RandomSampling} and \textit{RL-S2V}. Moreover, in targeted attacks, the transferability of attack strategy from \textit{RL-S2V} is also limited (see $\underline{italic}$ results) when comparing with \textit{RandomSampling}.}


\subsection{Other Transferability Evaluations}
\label{sec:transferability_data_model}

\subsubsection{Transferability on Attack Budget}
\hlt{Figure \ref{fig:attacks_different_k} shows the attack performance of our method on seen samples under different attack budgets. The performance curves illustrate our method learned effective and transferable attack strategies in untargeted and targeted attacks. Moreover, Figure \ref{fig:attacks_different_k} also empirically demonstrates that our method satisfies the \textit{operation ranking} principle in the evaluation framework since smaller budgets own larger marginal attack performance.}
\subsubsection{Transferability on Victim Model}
To further explore the transferability, we use the adversarial samples generated from GCN/GAT model to attack the victim GAT/GCN model. The attack results are shown in Table \ref{tab:attack_gatgcn}, and these adversarial samples achieve better attack performance than \textit{RandomSampling}. It indicates these victim models used for the same task on the same dataset may make consistent mistakes to some extent. 

\subsubsection{Transferability on Data Domain}
This evaluation explores the transferability of the attack strategy learned on one dataset to attack the victim models built on other datasets. We visualized the attack results on the victim GCN model in Figure \ref{fig:crossdomaingcn}. Firstly, the relative attack performance's colour indicates that when the source attacker and the target model have the same dataset, it achieves the best attack performance. Secondly, since the graphs in all datasets are small molecules or come from bioinformatics, the result values in Figure \ref{fig:crossdomaingcn} show the domain similarity of these graphs brings non-negative attack performance. 

These results reveal the potential risk of employing a pre-training model with a freezing setting \cite{DBLP:conf/iclr/HuLGZLPL20} in the downstream tasks. In Figure \ref{fig:crossdomaingcn}, we observe that when attacking the NCI-H23H dataset using the attack strategy from the ENZYMES dataset, the accuracy performance drops 8.9\%, which is larger than that of $RandSampling$ (4.7\%) and \textit{GradArgmax} (8.6\%). Once the adversary knows the victim model $f(\theta)$ uses the pre-training model $f_{emb}$, they could train a classifier $f'(\phi)$ on graphs $G_{f'}$ that come from the same domain of the data $G_{f}$ in $f(\theta)$, then use the attack strategy learned on $f'(\phi)$ and the embedding of $G_{f}$ from $f_{emb}$ to attack the victim models.

\subsection{Visualization of Attack Patterns}

\begin{table}[ht]
\centering
\caption{Classification Accuracy of the Victim Model (\%) on BA-2Motifs.}
\vspace{-5pt}
\label{tab:attack_2motifs}
\begin{tabular}{c cccccc}
\toprule
& \multicolumn{3}{c}{current samples} & \multicolumn{3}{c}{unseen samples} \\
\midrule
k             & 1                  & 2                  & 3                 & 1                  & 2                 & 3                 \\
\midrule
Clean         & 99.63             & 99.63             & 99.63            & 100.00             & 100.00            & 100.00            \\
\midrule
Rand          & 78.85              & 54.35              & 50.09             & 81.00              & 55.35             & \textbf{50.00}             \\
\midrule
Grad          & 60.50              & 51.13              & 50.13             & -                  & -                 & -                 \\
\midrule
RL-S2V        & \textbf{50.00}              & \textbf{50.00}     & \textbf{50.00}    & \textbf{50.00} & \textbf{50.00}    & \textbf{50.00}    \\
\midrule
\textbf{Ours} & \textbf{50.00}     & \textbf{50.00}     & \textbf{50.00}    & \textbf{50.00}     & \textbf{50.00}    & \textbf{50.00} \\  
\bottomrule
\end{tabular}
\vspace{-5pt}
\end{table}

To reveal the attack strategies of the projective ranking method, we visualize the perturbations in adversarial samples on the BA-2Motifs dataset based on the GCN model. BA-2Motifs is a synthetic dataset in which two motifs (\textit{House} and \textit{Pentagon}) are attached on random base graphs. Table \ref{tab:attack_2motifs} shows the attack results of different methods. 

\begin{figure}[t!]
  \centering
  \includegraphics[width=\linewidth]{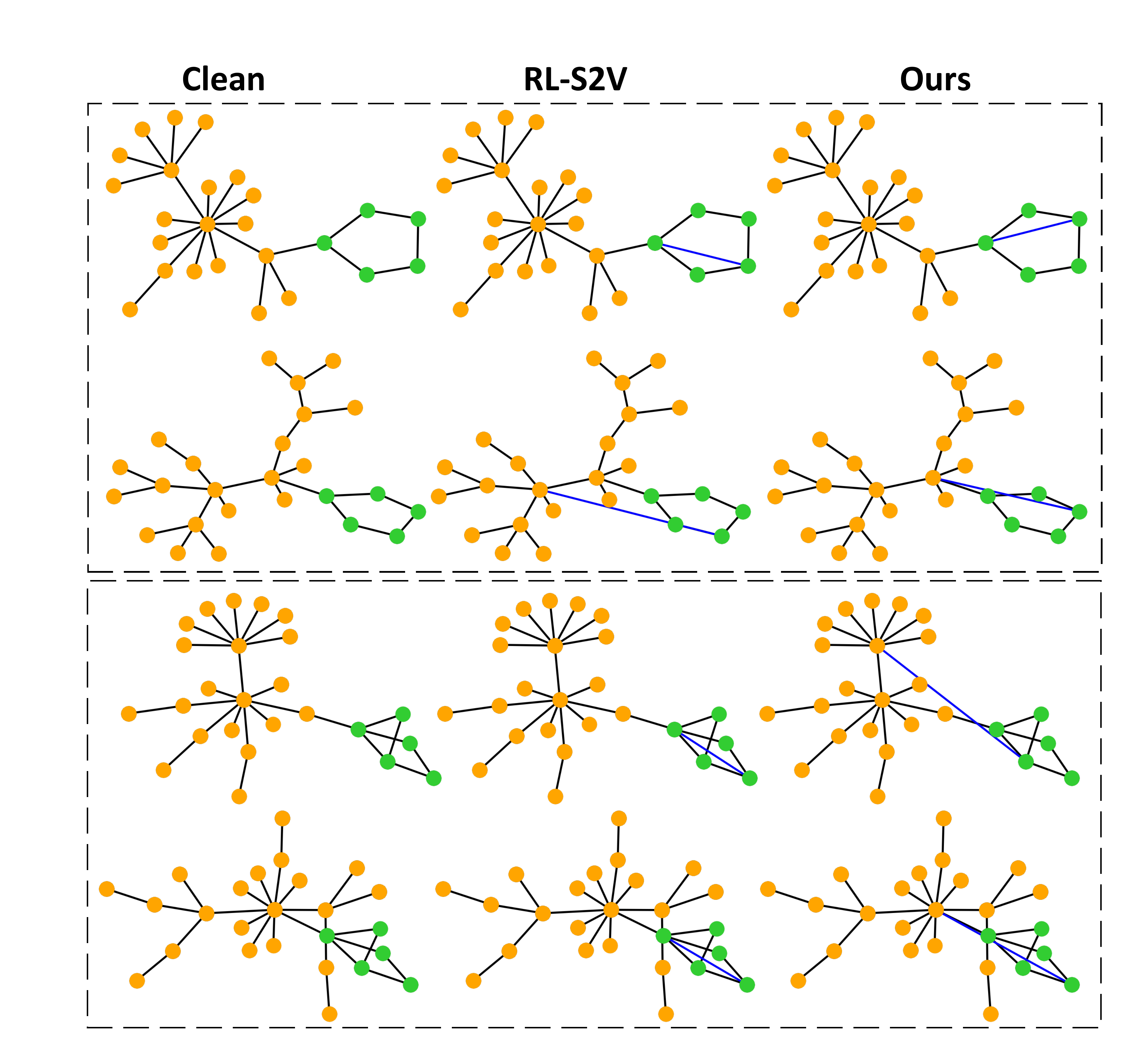}
  \vspace{-18pt}
  \caption{Visualization of adversarial samples. The graphs in the left column are clean samples, and the subgraphs in first/second block composed of green nodes are the motif "Pentagon"/"House". The graphs in the middle and right column are adversarial samples generated by \textit{RL-S2V} and our method, respectively. In this figure, the attackers attempt to fool the victim models to predict "House"/"Pentagon" on the adversarial samples in first/second block. Both \textit{RL-S2V} and our method present two attack patterns: \textbf{imitation of other motifs} and \textbf{collapse of self-motif}.}
  \label{fig:attackviz1}
  \vspace{-12pt}
\end{figure}

\begin{figure}[t!]
  \centering
  \includegraphics[width=\linewidth]{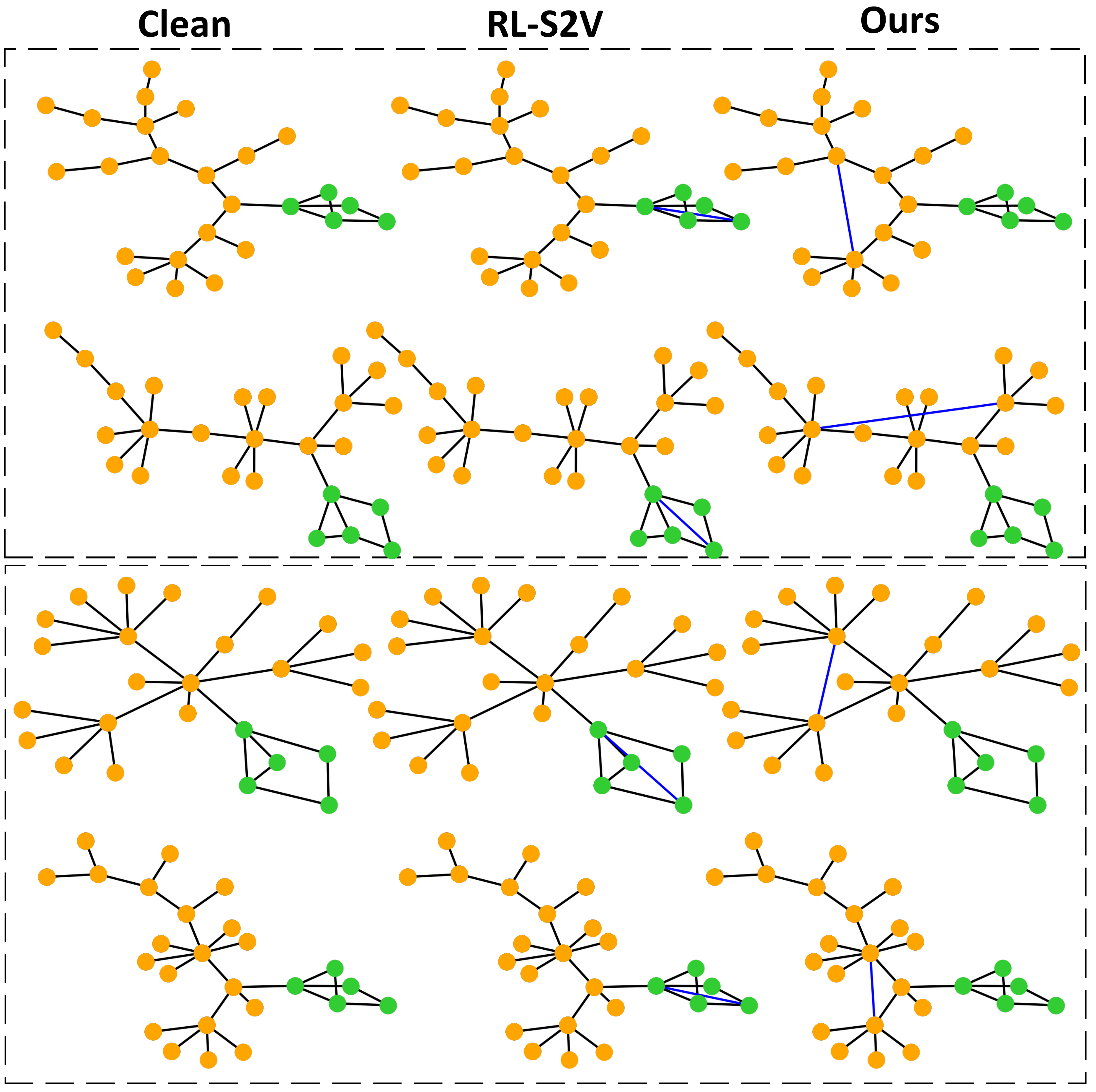}
  \vspace{-18pt}
  \caption{Visualization of adversarial samples.
  Compared with \textit{RL-S2V}, our method presents various attack patterns. They are the \textbf{coexistence of motifs} (first block) and the \textbf{fake self-motif} (second block).}
  \label{fig:attackviz2}
  \vspace{-18pt}
\end{figure}

In untargeted evasion attacks, we found two fundamental strategies in the generation of the adversarial samples. They are: (1) \textit{placing the adversarial samples in the high-confidence region of the other categories in the victim models}, and (2) \textit{moving the clean samples to the classification boundary or other low-confidence areas of the victim models}. We dive into the adversarial samples and observe some enlightening attack patterns in them:

\vspace{1mm}
\noindent
\textbf{Imitation of other motifs.} In the first row of the first block in Figure \ref{fig:attackviz1}, both \textit{RL-S2V} and our method add one edge in the \textit{Pentagon} so that there is a \textit{House} in the adversarial samples. In this way, the victim GCN model classifies the adversarial samples as \textit{House} graphs.

\vspace{1mm}
\noindent
\textbf{Collapse of self-motif.} In the second row of the first block and the second block in Figure \ref{fig:attackviz1}, both \textit{RL-S2V} and our method attack the victim model by adding one edge in which one node is in the self-motif. In this way, the attackers destroy the self-motif pattern by moving the samples to the low-confidence areas to fool the victim model. 

Although the attack results in Table \ref{tab:attack_2motifs} indicate \textit{RL-S2V} and our method have the same attack performance, we find the attack patterns of \textit{RL-S2V} are simple and lack variation from the second column in both Figure \ref{fig:attackviz1} and \ref{fig:attackviz2}. However, another two attack patterns are found in our method:

\vspace{0.5mm}
\noindent
\textbf{Coexistence of motifs.} In the first block of Figure \ref{fig:attackviz2}, our method shows a new attack pattern. One edge is added between two nodes that are outside of the self-motif in the clean samples. The exciting part of this edge is that the distance of the two end-nodes of it is 5, and after this perturbation operation, there is a motif \textit{Pentagon} in these samples. The coexistence of motifs is regarded as one behaviour that belongs to the second strategy in generating adversarial samples, in which the clean samples are moved to the decision boundary of the victim models.

\vspace{0.5mm}
\noindent
\textbf{Fake self-motif.} In the second block  of Figure \ref{fig:attackviz2}, our method adds one edge between two nodes outside of the self-motif, resulting in one triangle constructed in the adversarial samples. As a critical part of the \textit{House} motif, the constructed triangle could confuse the victim models since two ``similar" self-motifs exist in current samples. In this way, the samples are moved from the high-confidence region of the victim model to the low-confidence area.

In this section, we visualize the adversarial samples and summarize two basic attack strategies in them. We also found four attack patterns---imitation of other motifs, the collapse of self-motif, coexistence of motifs, and fake self-motif---of the adversarial samples. Compared with the \textit{RL-S2V}, the adversarial samples generated by the projective ranking method show various attack patterns. These attack patterns are helpful for the adversary to find the graph pattern differences between different data classes. Based on these observations, the attackers can customize the pattern operation to obtain adversarial samples of the specified type of clean graph samples.

\section{Conclusion}
\label{conclusion}
In this paper, we present the projective ranking approach to perform an evasion attack for graph classification. We contend that current evasion approaches either do not provide adequate attack performance without considering the perturbations' long-term benefits or require attackers to readjust the attack strategies when the attack environment changes. To that purpose, we define the perturbation space and propose an evaluation framework on the evasion attacks for graph classification. Then we first relax perturbation space for ranking its elements based on mutual information, and then we project the ranking into generating adversarial samples with a specified budget. The experimental results show that our method performs well in attack performance, and the learned attack strategies can be directly transferred to generate adversarial samples when the budget changes. Furthermore, the visualization of adversarial samples generated by our method shows a variety of attack patterns, which helps identify the vulnerability of the victim models.

Our future works include simultaneously making perturbations in structure and node features and performing evasion attacks on directed graphs and graphs with edge attributes. 
\hltt{In addition, two promising directions are exploring how imbalanced datasets influence the performance/robustness of GNNs and the interaction between robustness and explainability \cite{zhang2022trustworthy} by diving into adversarial samples.}

\section*{Acknowledgement}
This research was supported in part by an ARC Future Fellowship (FT210100097) and the National Natural Science Foundation of China (No. 61872360), and the CAS Project for Young Scientists in Basic Research (No. YSBR-008).

\ifCLASSOPTIONcaptionsoff
  \newpage
\fi
\bibliographystyle{IEEEtran}
\bibliography{reference_shortterm}

\appendices



\section{Challenges in Evasion Attack}
\label{challenges}
Unlike attack methods for node classification and link prediction, we find some challenges in designing evasion attack methods for graph classification. They are:\\
\textbf{(1) Huge search space.} The perturbation space is huge when the size of graph data is big. For example, it is hard for attackers to obtain the optimal adversarial samples when making edge perturbations. Different from mainly considering the link situation of the target node in the node-level task, the time complexity in the graph-level attack tasks is \begin{math}O(n^2)\end{math}, where \begin{math} n \end{math} is the node number of the graph. When \begin{math} d \end{math}-dimension binary attributes are associated with edges \cite{hu2020strategies}, this complexity changes to worse \begin{math}O(2^d n^2)\end{math}. 
As shown in the study on scalable GNN attacks \cite{GeislerSSZBG21}, it is usually difficult for the adversary to solve this discrete combinatorial optimization problem.\\
\textbf{(2) Profits maximization.} The profits here include the direct and indirect profits. The direct profit is the attack benefit since the primary goal of attackers is to obtain adversarial samples with high attack performance. The indirect profits contain learning transferable strategies and defect mining. In the aspect of transferable strategies, attackers expect to learn strategies that help attack unseen samples or generate adversarial samples under a changed perturbation budget. In the aspect of defect mining, the adversary yearns for revealing the defects or weaknesses of the victim models by analyzing the patterns in adversarial samples. The weaknesses of the victim models inspire attackers in designing more powerful attack methods. Once the flaws of the victim models are found, they could customize the attack strategies for different types of data to improve the attack performance.

\section{Different Measures of Attack Benefit}
\label{discuss_importance}

In this section, we will discuss the measure differences between {\itshape sensitivity, long-term benefit}, and {\itshape importance} concerning the \textit{attack benefit} principle.

\begin{figure}[ht!]
  \centering
  \includegraphics[width=\linewidth]{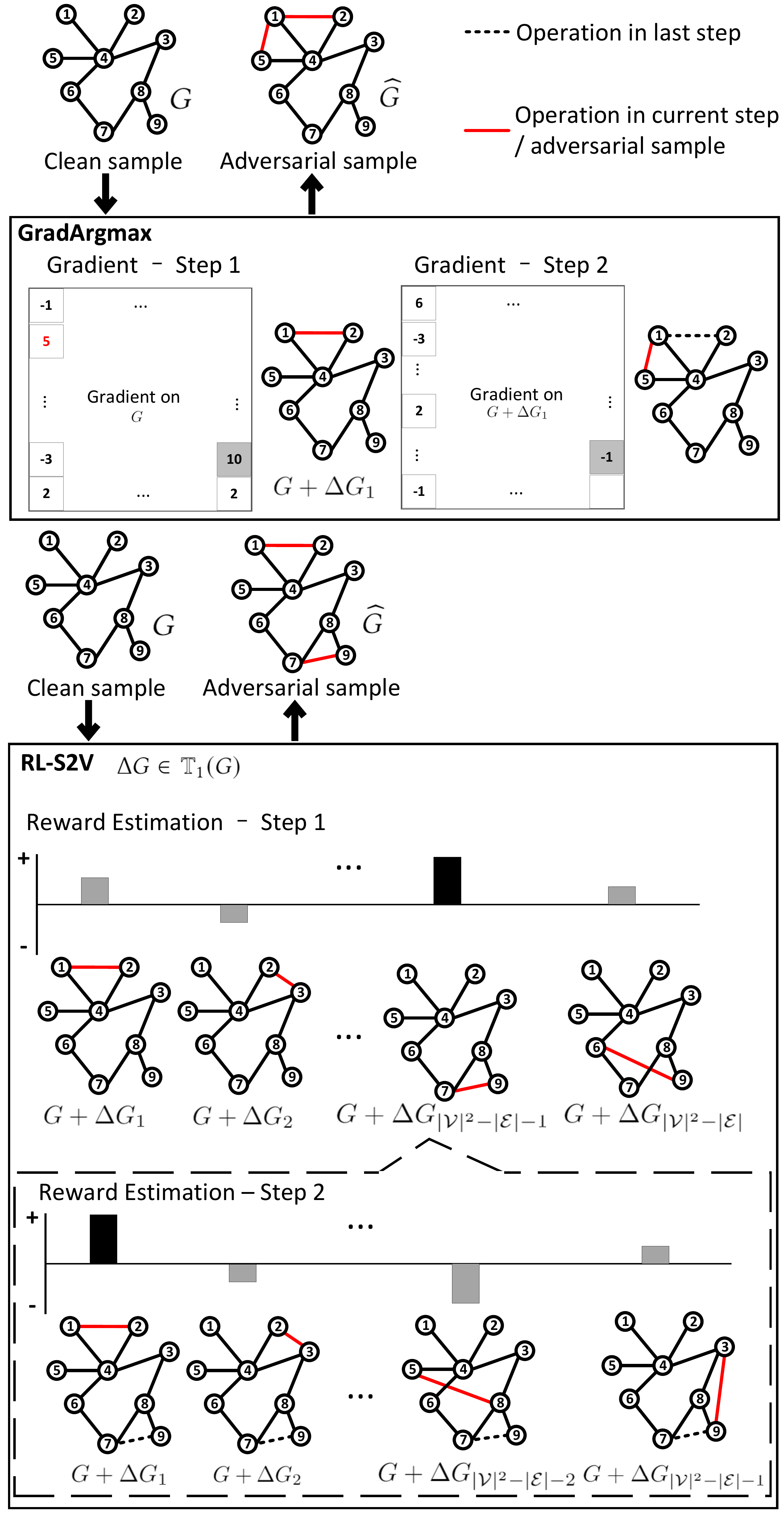}
  \caption{Illustration of \textit{GradArgmax} and \textit{RL-S2V}. \textit{GradArgmax} uses real-time gradient information to make perturbations, \textit{RL-S2V} utilizes the estimated long-term benefits to finish the attack.}
  \label{fig:rationalecomparison}
\end{figure}

\subsection{Sensitivity}
The sensitivity here refers to the gradient in the \textit{GradArgmax} method. It is intuitional to employ the gradient to make perturbation decision, since the gradient in \textit{GradArgmax} method exactly describe the structure or feature sensitivity of graph $G$ with respect to the victim model $f_{\theta}$. Taking the edge perturbation on graph $G$ as example, a coefficient $a_{u,v}$ is introduced for each nodes pair \begin{math}(u,v)\in \mathcal{V}\times\mathcal{V}\end{math}, and \begin{math}a_{u,v}=\mathbb{I}(v\in N(u))\end{math}. The message function in Section 3.2 changes to 
\begin{equation}
    m_{u}^{t+1}=\sum a_{u,v}M_{t}\left(h_u^t,h_v^t,e_{uv}\right).
\end{equation}
\hlt{To generate an adversarial sample, attackers first need to calculate the gradient on all $a_{u,v}$ 
\begin{equation}
    g_{u,v}=\frac{\partial \mathcal{L}}{\partial a_{u,v}},
\end{equation}
where $\mathcal{L}$ is the loss function of victim models.} Then \textit{GradArgmax} will choose the position with the largest positive $g_{u,v}$ to add an edge or the smallest negative $g_{u,v}$ to remove an edge. \textit{GradArgmax} will repeat the above process until it uses up all perturbation budget $k$. 

However, the gradient which measures the \textit{sensitivity} is not suitable enough to measure the \textbf{importance} of each perturbation operation for some reasons \cite{sundararajan2016gradients,sundararajan2017axiomatic,wuadversarial}:\\
\noindent
\textbf{(1)} The perturbation generated by \textit{GradArgmax} is sub-optimal. The principle in \textit{GradArgmax} is choosing the position with gradient extremum to make modifications. If there is an edge at the position with the largest positive gradient value or there is no edge at the position with the smallest negative gradient value, \textit{GradArgmax} will have to re-select the perturbation position.\\
\noindent
\textbf{(2)} The non-linear nature of GNN models limits the attack performance of \textit{GradArgmax}. In \textit{GradArgmax}, attackers expect to predict the change of loss function on every single operation with
\begin{equation}
\begin{split}
    B(\cdot\big|f_{\theta}) &= \mathcal{L}_k - \mathcal{L}_{k-1} =\Delta \mathcal{L} \\
    &=\mathcal{L}(G+\Delta G)-L(G)\\
    &\approx\ <\triangledown{G} \mathcal{L}(G),\Delta G>\\
    &=\sum g_{u,v}\Delta A_{u,v}, 
\end{split}
\end{equation}
which is only a first-order linear approximation function about $\Delta G$. However, to obtain powerful discrimination ability, GNN models usually contain nonlinear activation functions like ReLU. The linear approximation based on local gradient can't describe the change in the output of the victim models well when the input value changes greatly. Since the values in the adjacency matrix are discrete, the change from 0 to 1 or 1 to 0 may far exceeds the predictive ability of the above approximation. Taking \begin{math}y(x)=Relu(x)\end{math} as example, the gradient at $x=0$ is $g(0)=0$. When $x$ changes from $0$ to $1$, \(y(1)-y(0)\approx g(0)(1-0) = 0\) is far from the ground-truth $y(1)-y(0)=1$.\\
\noindent
\textbf{(3)} The iterative selection of perturbation operations based on local gradients cannot accurately describe the attacker's expectations. The {\itshape GradArgmax} attempts to use the local gradient to approximately measure the importance of each perturbation operation in the generation of adversarial samples. When the perturbation budget $k>$ 1, the local gradient is not a suitable approximate 
meassure. The reason is the local gradient $g_{u,v}$ at the $\kappa^{th}\ (1< \kappa \leq k)$ perturbation step is calculated based on current graph which is obtained by adding a perturbation graph $\Delta G \in \mathbb{T}_{\kappa-1}(G)$. Actually, in the $\kappa^{th}$ perturbation step, the attackers attempt to choose the perturbation operation which brings $\kappa^{th}$ largest attack benefit for the original clean sample $G$, but not the operation that brings $1^{st}$ largest benefit for current graph $G+\Delta G$ ($\Delta G \in \mathbb{T}_{\kappa-1}(G)$). So the gradient information is enough to predict the {\itshape sensitivity} but not a suitable meassure to identify the {\itshape importance} of each operation. 
\subsection{Long-term Benefit}
The {\itshape long-term benefit} here is the estimated reward of each operation in the adversarial samples generated by reinforcement learning. A conspicuous drawback of {\itshape GradArgmax} is that the greedy selection mechanism focuses on the short-term benefit (i.e., maximizing loss function) of the current operation, while neglect the benefit of current operation should be defined by its importance in final adversarial samples. In the training of {\itshape RL-S2V} with perturbation budget $k$, the estimated reward of each perturbation operation is 0 when the perturbation budget $k$ is not used up. Once the budget is exhausted, the adversarial samples obtain positive or negative reward defined by the output of the victim model $f_{\theta}$ on these samples. In the experience pool of {\itshape RL-S2V}, the \textit{long-term benefit} of each perturbation operation is determined by the final benefit of adversarial samples. In this way, {\itshape RL-S2V} can learn a more powerful attack strategy than {\itshape GradArgmax}.
\subsection{Importance}
The reward function in {\itshape RL-S2V} mainly focuses on the overall benefits caused by all perturbation operations in the adversarial samples. The {\itshape long-term benefit} of each perturbation operation in {\itshape RL-S2V} can only be evaluated by the reward function after the generation of adversarial examples exhausts all perturbation budget $k$. The {\itshape long-term benefit} reflects the \textit{absolute importance} of each perturbation operation according to the reward function under a specific budget $k$. This kind of absolute importance of the same perturbation operation is different under various perturbation budgets in attack methods based on reinforcement learning \cite{DBLP:conf/icml/DaiLTHWZS18,ma2019attacking}, so {\itshape RL-S2V} needs retraining to obtain attack strategy when the perturbation budget changes. From the perspective of \textit{operation ranking} principle, compared with the specific \textit{absolute importance} of each perturbation operation, attackers are more concerned about which perturbation operation should be selected from the perturbation space (i.e., \textit{relative importance}).

\end{document}